\documentclass[10pt,twocolumn,letterpaper]{article}

\usepackage[pagenumbers]{cvpr} % To force page numbers, e.g. for an arXiv version

\usepackage[utf8]{inputenc} % allow utf-8 input
\usepackage[T1]{fontenc}    % use 8-bit T1 fonts
\usepackage[dvipsnames]{xcolor}
\definecolor{cvprblue}{rgb}{0.21,0.49,0.74}
\usepackage[pagebackref,breaklinks,colorlinks,citecolor=cvprblue]{hyperref}
\usepackage{url}            % simple URL typesetting
\usepackage{booktabs}       % professional-quality tables
\usepackage{amsfonts}       % blackboard math symbols
\usepackage{nicefrac}       % compact symbols for 1/2, etc.
\usepackage{microtype}      % microtypography
\usepackage{amsmath,amssymb} % define this before the line numbering.
\usepackage{graphicx}
\usepackage{multirow}
\usepackage{multicol}
\usepackage{wrapfig}
\usepackage{caption}
\usepackage{subcaption}
\usepackage[ruled,vlined,linesnumbered]{algorithm2e}
\newcommand\PyCode[1]{\ttfamily\textcolor{black}{#1}}
\definecolor{commentcolor}{RGB}{110,154,155}
\newcommand\PyComment[1]{\ttfamily\textcolor{commentcolor}{\# #1}}

\def\paperID{6526} % *** Enter the Paper ID here
\def\confName{CVPR}
\def\confYear{2024}

\title{Unsupervised Discovery of Interpretable Directions \\ in h-space of Pre-trained Diffusion Models}

\author{
Zijian Zhang$^1$  \qquad  Luping Liu$^1$  \qquad  Zhijie Lin$^2$  \qquad  Yichen Zhu$^1$  \qquad Zhou Zhao$^1$\thanks{Corresponding author.}   \\
$^1$Department of Computer Science and Technology, Zhejiang University \\
$^2$ByteDance \\
{\tt\small ckczzj, luping.liu, linzhijie, yc\_zhu, zhaozhou@zju.edu.cn}  \\
}

\begin{document}
\maketitle
\begin{abstract}
We propose the first unsupervised and learning-based method to identify interpretable directions in h-space of pre-trained diffusion models.
Our method is derived from an existing technique that operates on the GAN latent space.
Specifically, we employ a shift control module that works on h-space of pre-trained diffusion models to manipulate a sample into a shifted version of itself, followed by a reconstructor to reproduce both the type and the strength of the manipulation.
By jointly optimizing them, the model will spontaneously discover disentangled and interpretable directions.
To prevent the discovery of meaningless and destructive directions, we employ a discriminator to maintain the fidelity of shifted sample.
Due to the iterative generative process of diffusion models, our training requires a substantial amount of GPU VRAM to store numerous intermediate tensors for back-propagating gradient.
To address this issue, we propose a general VRAM-efficient training algorithm based on gradient checkpointing technique to back-propagate any gradient through the whole generative process, with acceptable occupancy of VRAM and sacrifice of training efficiency.
Compared with existing related works on diffusion models, our method inherently identifies global and scalable directions, without necessitating any other complicated procedures.
Extensive experiments on various datasets demonstrate the effectiveness of our method.
\end{abstract}

\begin{figure}[t]
    \centering
    \includegraphics[width=0.45\textwidth]{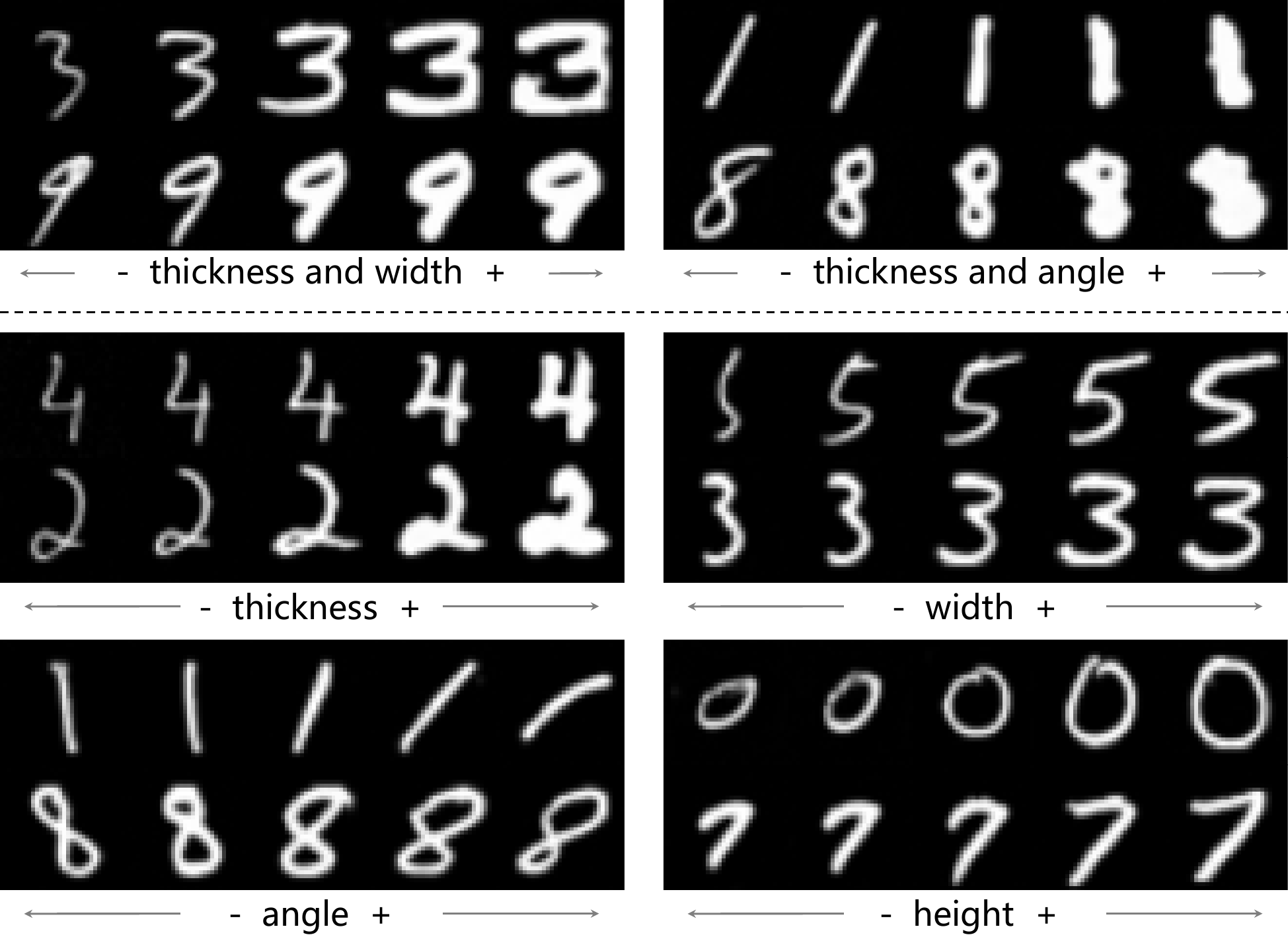}
    \caption{Interpretable directions discovered by our method in h-space of pre-trained DMs for MNIST. Each example contains five samples generated on the same discovered direction, but with different magnitudes. All attributes are manually interpreted.}
    \label{fig:mnist}
\end{figure}

\section{Introduction}
Recently, Diffusion Models (DMs)~\cite{feller1949theory,sohl2015deep,ho2020denoising} have exhibited remarkable capability to synthesize striking image samples.
% To control the generated images, one may rely on conditional DMs~\cite{rombach2022high,saharia2022photorealistic} or seek to adopt classifier-guided sampling technique~\cite{dhariwal2021diffusion}.
In the context of latent variable generative models, one can edit images by manipulating their corresponding latent variables, as the latent space of these models often have semantically meaningful directions.
% In the context of latent variable generative models, another alternative approach is to manipulate the corresponding latent variable of the image, as the latent spaces of these models often have semantically meaningful directions.
However, manipulating the latent variables of DMs directly may lead to distorted images or incorrect manipulation~\cite{kim2022diffusionclip}, as they lack high-level semantic information.
Some works~\cite{preechakul2021diffusion,zhang2022unsupervised} construct an external semantic latent space to address this issue.
Asyrp~\cite{kwon2022diffusion} explores the deepest bottleneck of the UNet as a local semantic latent space (h-space) to accommodate semantic image manipulation.

Despite their success, the discovery of meaningful directions in the semantic latent space of DMs relies on externel supervision, such as human annotation~\cite{preechakul2021diffusion,zhang2022unsupervised} and CLIP~\cite{radford2021learning, kwon2022diffusion}.
Many GAN-based works~\cite{voynov2020unsupervised, harkonen2020ganspace, shen2021closed, yuksel2021latentclr} have attempted to achieve analogous effects in an unsupervised manner, while there is a lack of related research on DMs.
A recent study~\cite{park2023unsupervised} proposes an unsupervised and training-free method to identify local semantic directions on the latent variable of certain individual image (i.e., $\mathbf{x}_{t}$).
% Specifically, the method adopts the Euclidean metric on h-space and leverages the connection beween h-space and thereproduce latent space to map the movement of h-space feature to that of corresponding latent variable.
However, the method cannot directly identify the global semantic directions for all samples, and fails to compute global semantic directions from the local ones for many interesting attributes.
Moreover, the method involves an iterative editing procedure with expensive Jacobian computations, known as geodesic shooting, to increase the editing strength or achieve multiple feature editing.

Inspired by the GANLatentDiscovery approach~\cite{voynov2020unsupervised} and the favorable properties of h-space, we propose the first unsupervised and learning-based method to directly identify global and scalable directions in the h-space of pre-trained DMs.
Figure~\ref{fig:mnist} shows a case of our method.
% In this paper, we follow the paradigm of GANLatentDiscovery approach~\cite{voynov2020unsupervised} and propose the first unsupervised and learning-based method to directly identify global and scalable directions in the h-space of pre-trained DMs.
Specifically, we first generate a sample from arbitrary Gaussian noise through regular DDIM reverse process.
Starting from the same noise, we then generate a shifted version of the sample through asymmetric DDIM reverse process controlled by a random shift direction index and a random magnitude.
Finally, a reconstructor is employed to reproduce both the index and the magnitude of the shift according to these two samples.
By jointly optimizing the shift control module and reconstructor, the model tends to discover disentangled and interpretable directions to make them easy to distinguish from each other.
Compared with the GAN-based counterpart~\cite{voynov2020unsupervised}, our DM-based method needs to address two primary issues.
The first one is how to maintain the fidelity of shifted samples, as the model can generate out-of-domain samples to simplify the reconstruction task.
\cite{voynov2020unsupervised} achieves this by regularizing the shifted latents because the distribution of GAN latent space is known.
However, the method is infeasible for us because the distribution of h-space is unknown.
Inspired by \cite{sehwag2022generating} and \cite{kim2022refining}, we introduce a discriminator to maintain the fidelity of shifted sample, preventing the discovery of meaningless and destructive directions.
The second one is how to train our model.
Unlike GANs that generate images in a single network pass, DMs generate images by iteratively denoising latents, which consume excessive VRAM to store intermediate tensors of each generative step for back-propagating gradient.
DiffusionCLIP~\cite{kim2022diffusionclip} proposes a GPU-efficient algorithm to alleviate this problem, but we find that it is ineffective for our method.
Therefore, we design a general VRAM-efficient algorithm based on gradient checkpointing technique~\cite{chen2016training, gruslys2016memory} to address the issue by back-propagating gradient node by node.
Compared with previous work~\cite{park2023unsupervised}, our method directly identifies global semantic directions, without necessitating to locate the local semantic directions with the same attributes from various individual samples and average them.
Moreover, our discovered semantic directions are inherently scalable, allowing us to scale the editing strength and achieve multiple feature editing easily.

\section{Related Work}
Our work focuses on the latent space of generative models.
Latent variable generative models such as GANs~\cite{goodfellow2014generative, karras2019style} and VAEs~\cite{kingma2013auto,rezende2014stochastic} inherently involve a latent space from which they generate data samples.
As an emerging latent variable generative model, DMs~\cite{feller1949theory,sohl2015deep,ho2020denoising} define their latent space on a sequence of corrupted data ($\mathbf{x}_{t}$) yielded through the forward process.
% Due to lack of high-level semantic information in this natural latent space, Diff-AE~\cite{preechakul2021diffusion} and PDAE~\cite{zhang2022unsupervised} explore DMs for representation learning via autoencoding, learning an extra semantic latent space from data.
% DisDiff~\cite{yang2023disdiff} designs a disentangling loss to force the encoder of PDAE to learn disentangled representation from data.
Asyrp~\cite{kwon2022diffusion} explores the deepest bottleneck of the UNet as a local semantic latent space (h-space) in frozen pre-trained DMs to accommodate semantic image manipulation.

The exploration of latent space have made remarkable advancements in recent years, particularly for GANs.
Many works~\cite{voynov2020unsupervised, harkonen2020ganspace, shen2021closed, yuksel2021latentclr} have made attempts to interpret and manipulate the latent space of GANs in an unsupervised fasion.
Recently, \cite{park2023unsupervised} has commenced the unsupervised exploration of the latent space of DMs to discover interpretable editing directions utilizing Riemannian geometry~\cite{arvanitidis2017latent, shao2018riemannian, chen2018metrics, arvanitidis2020geometrically}.
Specifically, the method adopts the Euclidean metric on h-space and identifies local semantic directions on the latent variable of certain individual sample (i.e., $\mathbf{x}_{t}$) that show large variability of the corresponding feature in h-space by employing the pullback metric~\cite{shao2018riemannian}.
The global semantic directions, which can be applied to all samples, are obtained by averaging the local semantic directions of individual samples.
To increase the editing strength and achieve multiple feature editing, an iterative editing procedure called geodesic shooting is employed to prevent the edited sample from escaping from the real data manifold.
Some normalization techniques are also employed to prevent the distortion due to editing.
% In contrast to the training-free method, we propose the first unsupervised and learning-based method to directly identify global and scalable directions.

\section{Background}
\subsection{Denoising Diffusion Probabilistic Models}
DDPMs~\cite{ho2020denoising} employ a forward Markov diffusion process $q(\mathbf{x}_{t} | \mathbf{x}_{t-1}) = \mathcal{N} (\mathbf{x}_{t}; \sqrt{1-\beta_{t}}\mathbf{x}_{t-1} , \beta_{t}\mathcal{I})$ to gradually convert the data distribution $q(\mathbf{x}_{0})$ to $\mathcal{N} (\mathbf{0}, \mathcal{I})$, where $\{\beta_{t}\}_{t=1}^{T}$ are some predefined variance schedule and $\{\mathbf{x}_{t}\}_{t=1}^{T}$ are latent variables of data $\mathbf{x}_{0}$.
The definitions enable us to directly sample $\mathbf{x}_{t}$ from $\mathbf{x}_{0}$ with $\mathbf{x}_{t} = \sqrt{\bar{\alpha}_{t}}\mathbf{x}_{0} + \sqrt{1-\bar{\alpha}_{t}}\epsilon$ for any $t$, where $\epsilon \sim \mathcal{N} (\mathbf{0}, \mathcal{I})$, $\alpha_{t}=1-\beta_{t}$ and $\bar{\alpha}_{t}=\prod_{i=1}^{t}\alpha_{i}$.
By approximating the reversal of forward process, the reverse process $p_{\theta}(\mathbf{x}_{0:T}) = p(\mathbf{x}_{T}) \prod_{t=1}^{T} p_{\theta}(\mathbf{x}_{t-1} | \mathbf{x}_{t})$ can generate samples starting from $p(\mathbf{x}_{T}) = \mathcal{N} (\mathbf{x}_{T}; \mathbf{0}, \mathcal{I})$ with the learned Gaussian transitions $p_{\theta}(\mathbf{x}_{t-1} | \mathbf{x}_{t}) = \mathcal{N} ( \mathbf{x}_{t-1}; \frac{1}{\sqrt{\alpha_{t}}} ( \mathbf{x}_{t} - \frac{1-\alpha_{t}}{\sqrt{1-\bar{\alpha}_{t}}} \epsilon_{t}^{\theta}(\mathbf{x}_{t}) ) , \frac{1-\bar{\alpha}_{t-1}}{1-\bar{\alpha}_{t}}\beta_{t}\mathcal{I} )$, where $\epsilon_{t}^{\theta}(\mathbf{x}_{t})$ is a function approximator that is trained to predict $\epsilon$ from $\mathbf{x}_{t}$.

\subsection{Denoising Diffusion Implicit Models}
DDIMs~\cite{song2020denoising} redefine the forward process of DDPMs as non-Markovian form, leading to a much more flexible reverse process to sample from.
Specifically, one can use some pre-trained $\epsilon_{t}^{\theta}(\mathbf{x}_{t})$ to sample $\mathbf{x}_{t-1}$ from $\mathbf{x}_{t}$ via $\mathbf{x}_{t-1} =  \sqrt{\bar{\alpha}_{t-1}} \mathbf{P}_{t}\big(\epsilon_{t}^{\theta}(\mathbf{x}_{t})\big)  + \mathbf{D}_{t}\big(\epsilon_{t}^{\theta}(\mathbf{x}_{t})\big) + \sigma_{t} \epsilon_{t}$, where $\mathbf{P}_{t}\big(\epsilon_{t}^{\theta}(\mathbf{x}_{t})\big) = \frac{\mathbf{x}_{t} - \sqrt{1 - \bar{\alpha}_{t}} \cdot \epsilon_{t}^{\theta}(\mathbf{x}_{t})}{\sqrt{\bar{\alpha}_{t}}}$ denotes the predicted $\mathbf{x}_{0}$ and $\mathbf{D}_{t}\big(\epsilon_{t}^{\theta}(\mathbf{x}_{t})\big) = \sqrt{1 - \bar{\alpha}_{t-1} - \sigma_{t}^{2}} \cdot \epsilon_{t}^{\theta}(\mathbf{x}_{t})$ denotes the direction pointing to $\mathbf{x}_{t}$.
$\epsilon_{t} \sim \mathcal{N}(\mathbf{0}, \mathcal{I})$ and $\sigma_{t}$ controls the stochasticity of forward process.
When $\sigma_{t}=0$ the process becomes deterministic.
The strides greater than $1$ are allowed for accelerated sampling.

\begin{figure*}[t]
    \centering
    \includegraphics[width=1.0\textwidth]{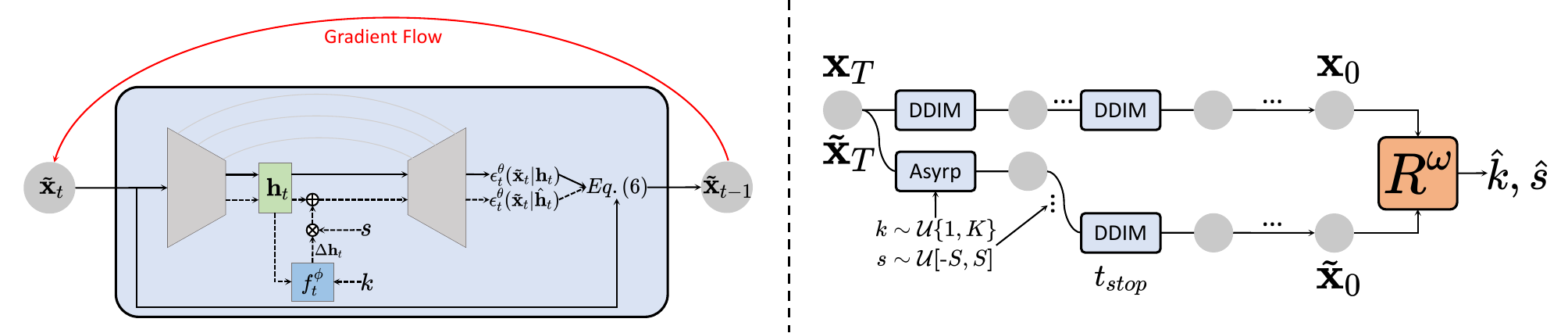}
    \caption{Left: One generation step of regular DDIM reverse process (solid line) and asymmetric DDIM reverse process (solid and dashed line). We take this one step as a unit for gradient checkpointing. Right: Conceptual illustration of our method. We generate $\mathbf{x}_{0}$ from $\mathbf{x}_{T}$ through DDIM reverse process. Starting from the same node ($\widetilde{\mathbf{x}}_{T} = \mathbf{x}_{T}$), we generate $\widetilde{\mathbf{x}}_{0}$ through asymmetric DDIM reverse process controlled by a random shift direction index $k$ and a random magnitude $s$, followed by DDIM reverse process after $t_{\text{stop}}$. A reconstructor $R^{\omega}$ is trained to reproduce $(k, s)$ from two generated samples.}
    \label{fig:model}
\end{figure*}

\subsection{h-space}
Asyrp~\cite{kwon2022diffusion} proposes asymmetric DDIM reverse process for semantic image manipulation, which explores the deepest bottleneck in the UNet of pre-trained DMs as semantic latent space known as h-space.
Specifically, it first inverts some real image $\mathbf{x}_{0}$ to $\mathbf{x}_{T}$ through deterministic DDIM forward process and then generates target $\widetilde{\mathbf{x}}_{0}$ starting from $\widetilde{\mathbf{x}}_{T} = \mathbf{x}_{T}$ through modified DDIM reverse process $p_{\theta}(\widetilde{\mathbf{x}}_{t-1} | \widetilde{\mathbf{x}}_{t})$:
\begin{equation} \label{asyrp}
    \left\{\begin{matrix}
        \begin{aligned}
            &\sqrt{\bar{\alpha}_{t-1}} \mathbf{P}_{t}\big(\epsilon_{t}^{\theta}(\widetilde{\mathbf{x}}_{t} | \hat{\mathbf{h}}_{t})\big)  + \mathbf{D}_{t}\big(\epsilon_{t}^{\theta}(\widetilde{\mathbf{x}}_{t} | \mathbf{h}_{t})\big) \\
            &\sqrt{\bar{\alpha}_{t-1}} \mathbf{P}_{t}\big(\epsilon_{t}^{\theta}(\widetilde{\mathbf{x}}_{t} | \mathbf{h}_{t})\big)  + \mathbf{D}_{t}\big(\epsilon_{t}^{\theta}(\widetilde{\mathbf{x}}_{t} | \mathbf{h}_{t})\big) \\
            &\sqrt{\bar{\alpha}_{t-1}} \mathbf{P}_{t}\big(\epsilon_{t}^{\theta}(\widetilde{\mathbf{x}}_{t} | \mathbf{h}_{t})\big)  + \mathbf{D}_{t}\big(\epsilon_{t}^{\theta}(\widetilde{\mathbf{x}}_{t} | \mathbf{h}_{t})\big) + \sigma_{t} \epsilon_{t}
        \end{aligned}
    \end{matrix}\right. ,
\end{equation}
where three formulas are used during $[T, t_{\text{edit}}]$, $[t_{\text{edit}}, t_{\text{noise}}]$ and $[t_{\text{noise}}, 0]$, respectively.
$\mathbf{h}_{t}$ is the h-space feature of $\widetilde{\mathbf{x}}_{t}$ and $\epsilon_{t}^{\theta}(\widetilde{\mathbf{x}}_{t} | \mathbf{h}_{t}) = \epsilon_{t}^{\theta}(\widetilde{\mathbf{x}}_{t})$.
$[T, t_{\text{edit}}]$ and $[t_{\text{noise}}, 0]$ are editing interval and quality boosting interval respectively.
DMs generate high-level context during editing interval~\cite{choi2022perception}, and we can achieve semantic changes by employing asymmetric DDIM reverse process within it.
The key idea of Asyrp is to replace original $\mathbf{h}_{t}$ with modified $\hat{\mathbf{h}}_{t} = \mathbf{h}_{t} + \Delta \mathbf{h}_{t}^{\text{attr}}$ for $\mathbf{P}_{t}(\cdot)$ during editing interval, which will shift the generative trajectory towards desired attribute.
A small neural network $f_{t}^{\phi}(\mathbf{h}_{t})$ with parameter $\phi$ is trained to predict $\Delta \mathbf{h}_{t}^{\text{attr}}$ by minimizing CLIP directional loss~\cite{radford2021learning, gal2022stylegan, kim2022diffusionclip}.

\section{Method}
\subsection{Overall Framework}
Our fundamental idea is based on seminal GANLatentDiscovery approach~\cite{voynov2020unsupervised}, which identifies interpretable directions in the latent space of a pre-trained GAN model without any form of supervision.
We aim to perform similar operations on the h-space~\cite{kwon2022diffusion} of some pre-trained DM to achieve analogous effects.
Figure~\ref{fig:model} shows the conceptual illustration of our method.
Specifically, we assume that there are $K$ interpretable directions to be discovered in the h-space.
To avoid discovering "abrupt" directions such as shifting all samples to some fixed pattern~\cite{voynov2020unsupervised}, we also assume that all directions are scalable within the range of $[-S, S]$.
During training, we first draw a $\mathbf{x}_{T} \sim \mathcal{N} (\mathbf{0}, \mathcal{I})$ as starting node and run $M$-step DDIM reverse process to generate $\mathbf{x}_{0}$.
Next we randomly select an integer $k \sim \mathcal{U}\{1, K\}$ as shift direction index and sample a scalar $s \sim \mathcal{U}[-S, S]$ as shift magnitude.
Staring from the same node $\widetilde{\mathbf{x}}_{T} = \mathbf{x}_{T}$, we then run $M$-step following reverse process to get shifted $\widetilde{\mathbf{x}}_{0}$:
\begin{equation}\label{shift_reverse_process}
    \widetilde{\mathbf{x}}_{t-1} =  \left\{\begin{matrix}
        \begin{aligned}
            &\sqrt{\bar{\alpha}_{t-1}} \mathbf{P}_{t}\big(\epsilon_{t}^{\theta}(\widetilde{\mathbf{x}}_{t} | \hat{\mathbf{h}}_{t})\big)  + \mathbf{D}_{t}\big(\epsilon_{t}^{\theta}(\widetilde{\mathbf{x}}_{t} | \mathbf{h}_{t})\big) \\
            &\sqrt{\bar{\alpha}_{t-1}} \mathbf{P}_{t}\big(\epsilon_{t}^{\theta}(\widetilde{\mathbf{x}}_{t} | \mathbf{h}_{t})\big)  + \mathbf{D}_{t}\big(\epsilon_{t}^{\theta}(\widetilde{\mathbf{x}}_{t} | \mathbf{h}_{t})\big)
        \end{aligned}
    \end{matrix}\right. ,
\end{equation}
where the first line is asymmetric DDIM reverse process~\cite{kwon2022diffusion} for $[T, t_{\text{stop}}]$ and the second line is regular DDIM reverse process for $[t_{\text{stop}}, 0]$.
A trainable shift block $f_{t}^{\phi}$ is employed to control this asymmetric DDIM reverse process according to $(k, s)$, and we will discuss it in Section~\ref{shiftblock}.
We refer to $[T, t_{\text{stop}}]$ as the shifting interval, similar to the editing interval in Eq.(\ref{asyrp}).
For simplicity, we don't employ quality boosting interval in Eq.(\ref{asyrp}).
Finally, the generated $\mathbf{x}_{0}$ and $\widetilde{\mathbf{x}}_{0}$ are passed to a reconstructor $R^{\omega}$ to reproduce $(k, s)$.
By jointly optimizing shift block and reconstructor, the model tends to discover disentangled and interpretable directions to make them easy to distinguish from each other.
$K$, $S$, $M$ and $t_{\text{stop}}$ are all hyperparameters, and all components of pre-trained DM are frozen during training.
We then describe further details in the following sections.

\begin{figure}[t]
    \centering
    \includegraphics[width=0.45\textwidth]{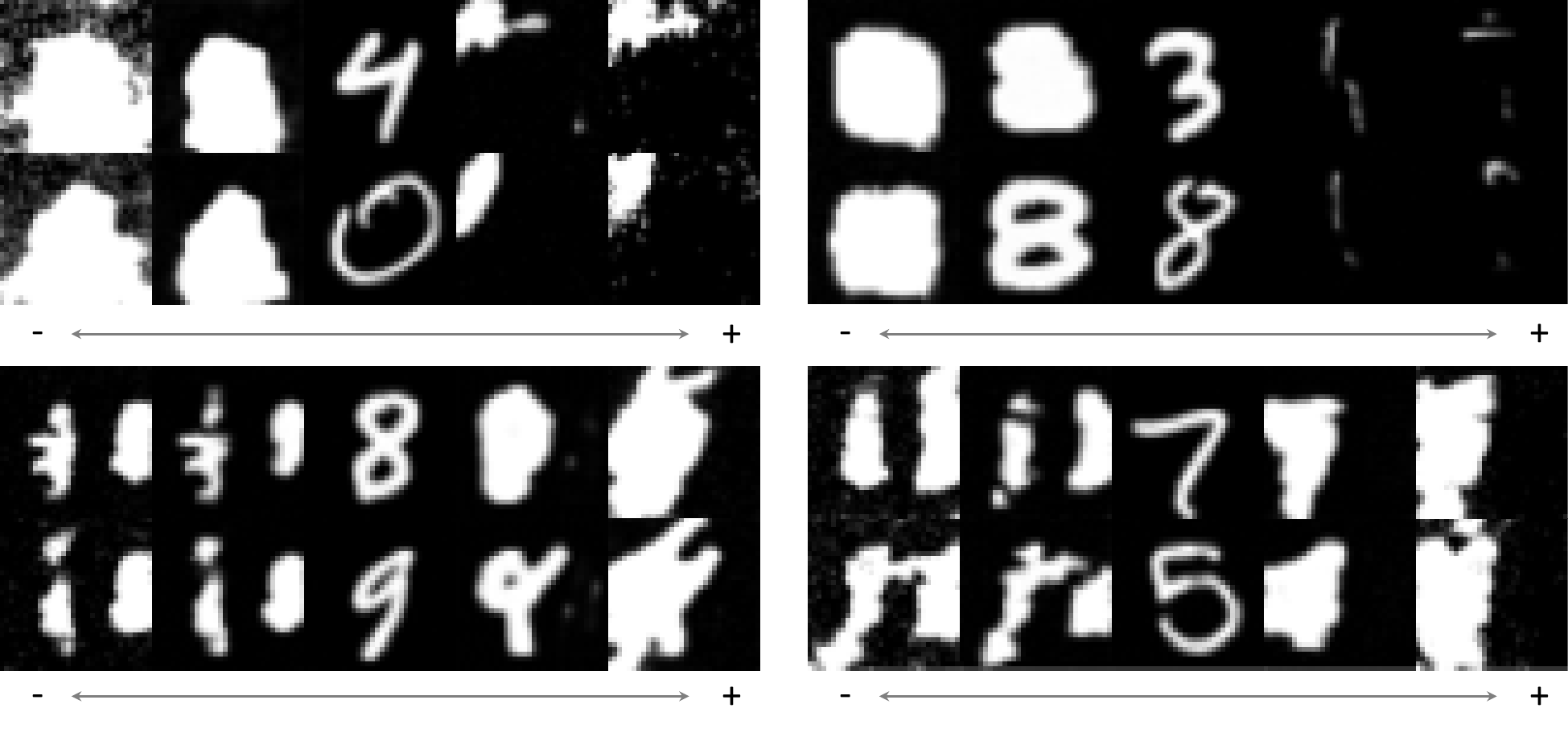}
    % \caption{Three meaningless directions discovered by our method in the h-space of DM pre-trained on MNIST when only using weighted $l_{1}$ loss.}
    \caption{When only using weighted $l_{1}$ loss, our method fails to identify meaningful directions in the h-space of pre-trained DM on MNIST. The model shifts all samples to some fixed and meaningless pattern to simplify the reconstruction tasks.}
    \label{fig:fail}
\end{figure}

\subsection{Shift Block}\label{shiftblock}
Asyrp~\cite{kwon2022diffusion} trains a small time-dependent neural network with two $1 \times 1$ convolutions (one input convolution and one output convolution) to predict an attribute-related shift $\Delta \mathbf{h}_{t}^{\text{attr}}$ given corresponding h-space feature $\mathbf{h}_{t}$.
We employ the same implementation, but with $K$ independent $1 \times 1$ output convolutions, as our shift block $f_{t}^{\phi}$.
For each shift direction index $k$, we use the $k$-th output convolution to predict $\Delta \mathbf{h}_{t}^{k} = f_{t}^{\phi}(\mathbf{h}_{t}, k)$.
For shift magnitude $s$, we leverage the linearity property of h-space (i.e., linearly scaling a $\Delta \mathbf{h}_{t}^{\text{attr}}$ reflects corresponding amount of attibute change in samples) and directly multiply the predicted $\Delta \mathbf{h}_{t}^{k}$ by $s$.
We replace original h-space feature $\mathbf{h}_{t}$ with shifted one $\hat{\mathbf{h}}_{t} = \mathbf{h}_{t} + s \cdot \Delta \mathbf{h}_{t}^{k}$ for the asymmetric DDIM reverse process during $[T, t_{\text{stop}}]$ in Eq.(\ref{shift_reverse_process}).

\subsection{Maintaining the Fidelity of Shifted Samples}
% How to maintain the fidelity of shifted sample is a main concern for us, as the model can shift all samples to some out-of-domain pattern to simplify the reconstruction of $(k, s)$.
How to maintain the fidelity of shifted samples is a primary concern for us, as the model can steer the reverse process off data-manifold and generate out-of-domain samples to simplify the reconstruction of $(k, s)$.
GANLatentDiscovery~\cite{voynov2020unsupervised} carefully chooses the initialization of shift direction vectors (unit length or orthonormal) to make the shifted latents still lie in the latent space, ensuring that the shifted samples remain within the data distribution.
However, h-space is a high-level semantic latent space whose distribution remains unknown to us, thus it is not feasible to achieve this by regularizing the predicted $\Delta \mathbf{h}_{t}^{k} = f_{t}^{\phi}(\mathbf{h}_{t}, k)$.
DiffusionCLIP~\cite{kim2022diffusionclip} and Asyrp~\cite{kwon2022diffusion} employ $l_{1}$ loss to prevent unwanted changes and preserve the identity of the object in the edited samples.
Likewise, we originally worked with a weighted $l_{1}$ loss $\frac{\gamma}{\gamma+|s|} \cdot || \widetilde{\mathbf{x}}_{0} - \mathbf{x}_{0} ||_{1}$ to to prevent unwanted changes and encourage more changes for larger $|s|$, where $\gamma$ is a constant.
However, we find that it is hard to balance this loss and reconstruction loss, and the model still struggles to maintain the fidelity of shifted samples, learning  meaningless and destructive directions as shown in Figure~\ref{fig:fail}.
We attribute this to lack of explicit supervision, such as CLIP directional loss, and $l_{1}$ loss only serves as a form of spatial regularization.
Therefore, we need an high-level regularization to help maintain the fidelity of shifted samples.

% Inspired by \cite{sehwag2022generating} and \cite{kim2022refining}, which train a discriminator to distinguish between real data and fake samples generated by a pre-trained DM and use it to guide the sampling process, we also introduce a discriminator $D^{\psi}$ for our method.

\cite{sehwag2022generating} and \cite{kim2022refining} have recently introduced a time-dependent discriminator for pre-trained DMs that distinguishes between the noisy version of real data and generated samples to guide the sampling process, enforcing generated samples to stay close to the real data manifold.
Taking inspiration from them, we also introduce a discriminator into our method.
% Inspired by \cite{sehwag2022generating} and \cite{kim2022refining}, which train a time-dependent discriminator that distinguishes between the noisy version of real data and fake samples generated by a pre-trained DM, and employ it to guide the sampling process to enforce synthetic samples to stay close to the real data manifold, we also introduce a discriminator for our method.
Instead of a time-dependent discriminator, we find that a discriminator that only works on clean data is enough.
Specifically, we consider $\mathbf{x}_{0}$ as real samples and $\widetilde{\mathbf{x}}_{0}$ as fake samples, and train a discriminator $D^{\psi}$ to distinguish them.
When training the shift block, we force it to generate samples that can deceive $D^{\psi}$.
Note that we initialize $K$ output convolutions of shift block with zeros to ensure that $\mathbf{x}_{0}$ and $\widetilde{\mathbf{x}}_{0}$ are identical at the begining of training~\cite{zhang2023adding}.
In this way, we find that the training reaches a stationary Nash equilibrium from the beginning and almost maintains it throughout.

\subsection{VRAM-efficient Training Algorithm}\label{sec:algorithm}
Unlike GANs and VAEs, DMs generate samples by iteratively denoising latents $\mathbf{x}_{t}$ with the same network, which makes our optimization similar to the process of training a recursive neural network~\cite{rumelhart1985learning, kim2022diffusionclip}.
This leads to a heavy usage of GPU VRAM, as GPU has to keep the intermediate tensors of every generation step in VRAM to back-propagate gradient.
% We refer to this as the vanilla algorithm.
To alleviate this problem, DiffusionCLIP~\cite{kim2022diffusionclip} and Asyrp~\cite{kwon2022diffusion} adopt a GPU-efficient algorithm that calculates loss using predicted $\mathbf{x}_{0}$ (i.e., $\frac{\mathbf{x}_{t} - \sqrt{1 - \bar{\alpha}_{t}} \cdot \epsilon_{t}^{\theta}(\mathbf{x}_{t})}{\sqrt{\bar{\alpha}_{t}}}$) and performs optimization at every timestep $t$ of reverse process.
However, we find the algorithm doesn't work for our method.
One likely reason is also that we lack explicit supervision, such as CLIP directional loss, which makes the optimizations at different timesteps inconsistent.

In light of this, to save VRAM, we propose a VRAM-efficient training algorithm based on gradient checkpointing technique~\cite{chen2016training, gruslys2016memory} that only needs to keep the intermediate tensors of one generation step in VRAM during training.
Specifically, as shown in Figure~\ref{fig:model}, one generation step can be seen as a transformation from one node $\widetilde{\mathbf{x}}_{t}$ to its succeeding node $\widetilde{\mathbf{x}}_{t-1}$, and the reverse process can be seen as a chain connected by sequential nodes ($\widetilde{\mathbf{x}}_{T} \rightarrow \widetilde{\mathbf{x}}_{T-1} \rightarrow \cdots \rightarrow \widetilde{\mathbf{x}}_{0}$).
The gradient flows on this chain in reverse.
% The gradient flows on this chain in reverse, from one node $\widetilde{\mathbf{x}}_{t-1}$ to its ancestral node $\widetilde{\mathbf{x}}_{t}$.
% According to chain rule $\frac{\partial \mathcal{L}}{\partial \widetilde{\mathbf{x}}_{t}} = \frac{\partial \mathcal{L}}{\partial \widetilde{\mathbf{x}}_{t-1}} \frac{\partial \widetilde{\mathbf{x}}_{t-1}}{\partial \widetilde{\mathbf{x}}_{t}}$, the gradient of node $\widetilde{\mathbf{x}}_{t}$ is only related to the gradient of node $\widetilde{\mathbf{x}}_{t-1}$ and 
% As long as we know the ancestral node $\widetilde{\mathbf{x}}_{t}$ of current node $\widetilde{\mathbf{x}}_{t-1}$, we can propagate gradient node by node, without need for the intermediate tensors among previous nodes $\widetilde{\mathbf{x}}_{<t}$.
% The gradient flows on this chain in reverse, so we can propagate gradient node by node, from current node $\widetilde{\mathbf{x}}_{t-1}$ to its ancestral node $\widetilde{\mathbf{x}}_{t}$, without need for the intermediate tensors among previous nodes $\widetilde{\mathbf{x}}_{>t}$.
Rather than back-propagate gradient through all nodes at one time (i.e., from node $\widetilde{\mathbf{x}}_{0}$ to node $\widetilde{\mathbf{x}}_{T}$), inspired by gradient checkpointing technique~\cite{chen2016training, gruslys2016memory}, we can back-propagate gradient node by node, from current node $\widetilde{\mathbf{x}}_{t-1}$ to its ancestral node $\widetilde{\mathbf{x}}_{t}$.
This needs us to run the reverse process twice.
Specifically, we first run the reverse process in the gradient-disabled mode and record all intermediate nodes ($\widetilde{\mathbf{x}}_{T}, \widetilde{\mathbf{x}}_{T-1}, \cdots, \widetilde{\mathbf{x}}_{1}$).
Then we run the reverse process in the gradient-enabled mode for the second time, but in reverse order ($\widetilde{\mathbf{x}}_{1} \rightarrow \widetilde{\mathbf{x}}_{0}, \widetilde{\mathbf{x}}_{2} \rightarrow \widetilde{\mathbf{x}}_{1}, \cdots, \widetilde{\mathbf{x}}_{T}\rightarrow \widetilde{\mathbf{x}}_{T-1}$).
For the first step ($\widetilde{\mathbf{x}}_{1} \rightarrow \widetilde{\mathbf{x}}_{0}$), we use the recorded node $\widetilde{\mathbf{x}}_{1}$ as input to generate succeeding node $\widetilde{\mathbf{x}}_{0}$, compute loss $\mathcal{L}$ using $\widetilde{\mathbf{x}}_{0}$, calculate $\frac{\partial \mathcal{L}}{\partial \widetilde{\mathbf{x}}_{0}}$ and $\frac{\partial \widetilde{\mathbf{x}}_{0}}{\partial \widetilde{\mathbf{x}}_{1}}$, get $\frac{\partial \mathcal{L}}{\partial \widetilde{\mathbf{x}}_{1}} = \frac{\partial \mathcal{L}}{\partial \widetilde{\mathbf{x}}_{0}} \frac{\partial \widetilde{\mathbf{x}}_{0}}{\partial \widetilde{\mathbf{x}}_{1}}$ and cache $\frac{\partial \mathcal{L}}{\partial \widetilde{\mathbf{x}}_{1}}$.
For each subsequent step, we use the recorded node $\widetilde{\mathbf{x}}_{t}$ as input to generate succeeding node $\widetilde{\mathbf{x}}_{t-1}$, calculate $\frac{\partial \widetilde{\mathbf{x}}_{t-1}}{\partial \widetilde{\mathbf{x}}_{t}}$, multiply $\frac{\partial \widetilde{\mathbf{x}}_{t-1}}{\partial \widetilde{\mathbf{x}}_{t}}$ by cached $\frac{\partial \mathcal{L}}{\partial \widetilde{\mathbf{x}}_{t-1}}$, get $\frac{\partial \mathcal{L}}{\partial \widetilde{\mathbf{x}}_{t}} = \frac{\partial \mathcal{L}}{\partial \widetilde{\mathbf{x}}_{t-1}} \frac{\partial \widetilde{\mathbf{x}}_{t-1}}{\partial \widetilde{\mathbf{x}}_{t}}$ and cache $\frac{\partial \mathcal{L}}{\partial \widetilde{\mathbf{x}}_{t}}$.
The gradient will be accumulated to shift block if the step is asymmetric.
In this way, GPU only processes one generation step at any time during back-propagating gradient on this chain.
We put detailed procedure of this algorithm in Appendix A.
More generally, the algorithm can help to back-propagate any gradient through the whole generative process.
This enable us to train or fine-tune the diffusion models according to any objective involving generated sample $\widetilde{\mathbf{x}}_{0}$, such as CLIP directional loss~\cite{radford2021learning, gal2022stylegan, kim2022diffusionclip} and human reward feedback~\cite{lee2023aligning}.
We will benchmark its performance in Section~\ref{sec:benchmark}.

\section{Experiments}
We evaluate our method on MNIST~\cite{lecun1998gradient}, AnimeFaces~\cite{jin2017towards}, CelebAHQ~\cite{karras2017progressive} and AFHQ-dog~\cite{choi2018stargan} datasets.
% We pre-train DMs on these datasets using ADM architecture~\cite{dhariwal2021diffusion}.
All experiments are conducted in a completely unsupervised manner, thus all attributes are manually interpreted.
For brevity, we use the notation such as "MNIST32-400-20-32-5" to name our model, which means that we use a DM pre-trained on $32 \times 32$ MNIST dataset to perform our method with $t_{\text{stop}}=400, M=20, K=32, S=5$.
We put all implementation details in Appendix B, including network architecture, hyperparameters and training configurations.
More samples of following experiments can be found in Appendix C.

\subsection{Benchmark of VRAM-efficient Algorithm}\label{sec:benchmark}
\begin{figure}[t]
    \centering
    \includegraphics[width=0.45\textwidth]{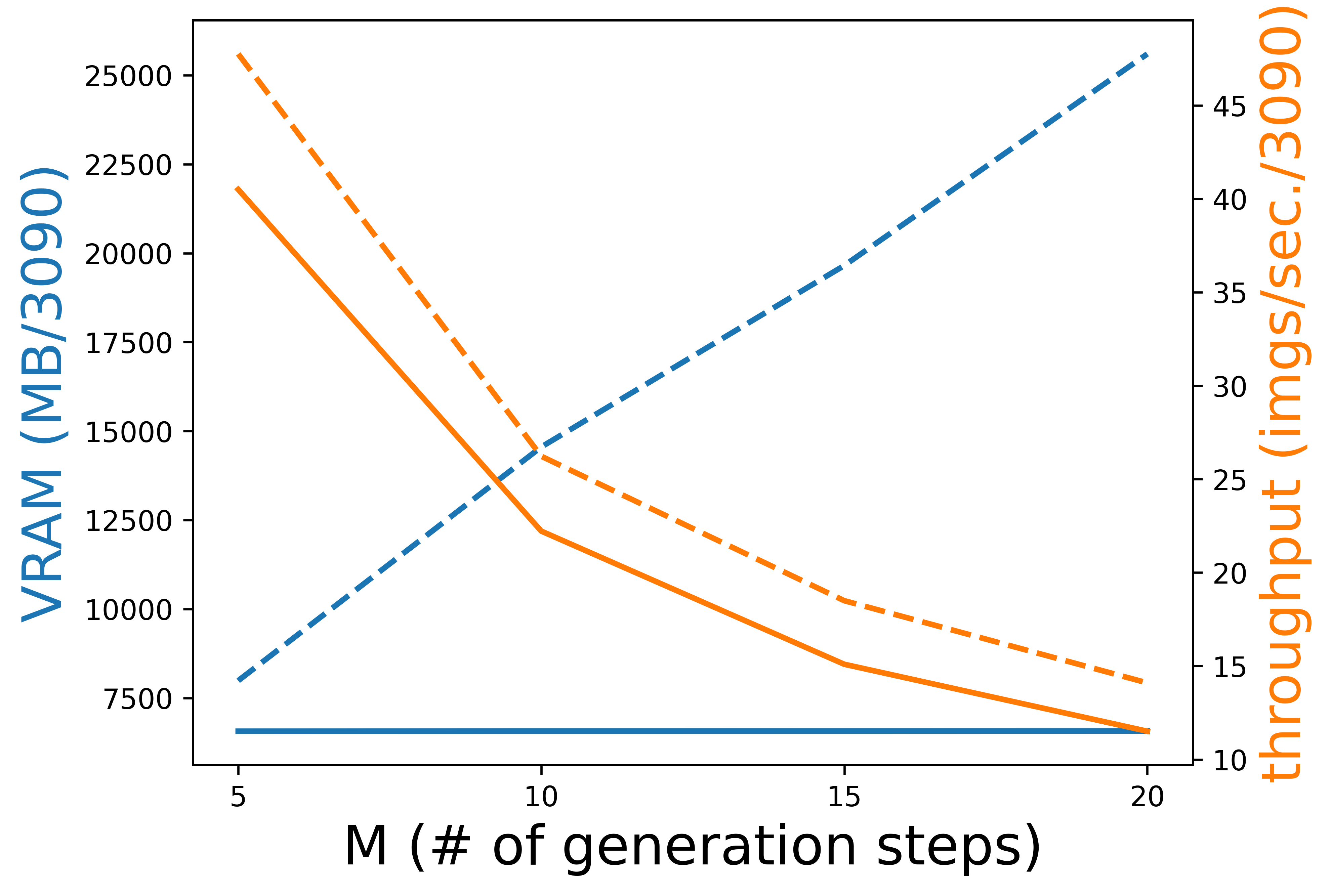}
    \caption{The comparative results of our algorithm (solid line) and the vanilla one (dashed line).}
    \label{fig:benchmark}
\end{figure}

We demonstrate better VRAM usage and comparable training efficiency of our algorithm compared with the vanilla one (i.e., no use of gradient checkpointing technique).
Note that the two algorithms have completely identical training effects.
Specifically, we conduct training using both algorithms for "MNIST32-400-$M$-32-5" with varying values of $M$.
We use $4$ Nvidia RTX 3090 GPUs for distributed training and set the batch size to $128$ ($32$ for each 3090) to inspect their training VRAM (in MB/3090) and throughput (in imgs/sec./3090).
Figure~\ref{fig:benchmark} show the comparative results.
As can be seen, our algorithm maintains consistent VRAM usage as $M$ increases, while the vanilla one consumes progressively more VRAM.
The VRAM usage of vanilla algorithm for $M=20$ has exceeded the maximum VRAM of Nvidia RTX 3090 (24GB).
Moreover, our algorithm achieves comparable throughput to the vanilla one, despite requiring an additional run of the reverse process in gradient-disabled mode.
Note that the benchmark is conducted on the small "MNIST32-400-$M$-32-5" model, and the VRAM usage will become extremely unacceptable when using larger pre-trained DMs, especially for large $M$.
Our algorithm is considerably more practical in comparison to the vanilla one.

As a comparison, a recent work ChatFace~\cite{yue2023chatface} adopts the vanilla algorithm to train a 4-layer MLP to predict manipulation direction for the latent space of Diff-AE~\cite{preechakul2021diffusion} using CLIP directional loss.
Although they have 8 Nvidia 3090 GPUs, they can only set $M=8$ with a batch size of 8 (i.e., only one training sample for each GPU).
The VRAM-efficient algorithm, however, can handle this problem, enabling more sampling steps for better sample quality and larger batch size for more stable training.

\subsection{Interpretable Directions}
We present qualitative examples induced by interpretable directions identified with our method in this section.
For each example, we generate five images staring from the same $\widetilde{\mathbf{x}}_{T} \sim \mathcal{N} (\mathbf{0}, \mathcal{I})$ with Eq.(\ref{shift_reverse_process}), while keeping the same $k$ and varying $s$ uniformly from $-S$ to $S$.
Therefore, the middle one represents the original image.
Figure~\ref{fig:mnist} shows the interpretable directions discovered by "MNIST32-400-20-32-5".
In our experiments, we usually set a sufficiently large value for $K$, surpassing the potential number of disentangled interpretable directions inherent in the data.
Consequently, our method also discovers many entangled directions, such as "thickness and width" for MNIST, which can be considered as a combination of "thickness" and "width".
Therefore, we only report approximatively disentangled directions in following examples.
Figure~\ref{fig:anime}~\ref{fig:celebahq}~\ref{fig:afhq} show the interpretable directions discovered by "Anime64-400-40-128-2", "CelebAHQ128-500-40-256-2" and "AFHQ128-400-40-256-2", respectively.
As we can see, our method can identify numerous meaningful semantic directions that can be generally applied to all samples and smoothly transit all samples along them.
Furthermore, our method can also handle the situation where the original image is positioned at different locations along a direction.
% We also evaluate our method in terms of quantitative results, which are put in Appendix C due to limited space.

\begin{figure*}[t]
    \centering
    \includegraphics[width=0.95\textwidth]{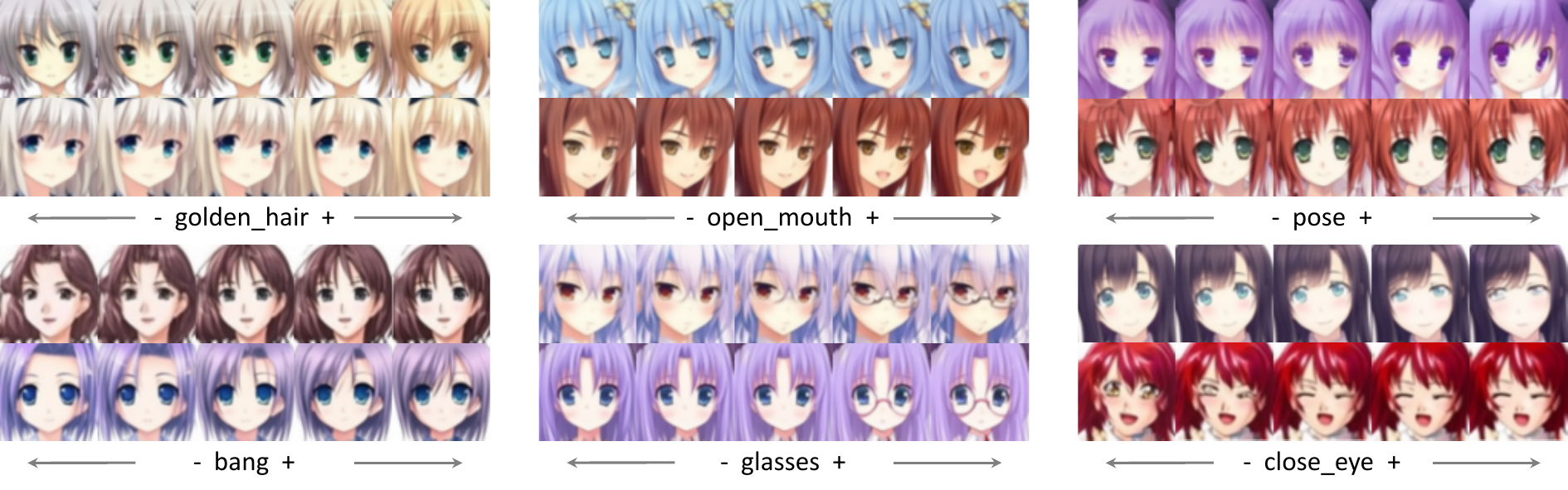}
    \vspace{-0.3cm}
    \caption{Interpretable directions discovered by "Anime64-400-40-128-2".}
    \label{fig:anime}
\end{figure*}

\begin{figure*}[t]
    \centering
    \includegraphics[width=0.95\textwidth]{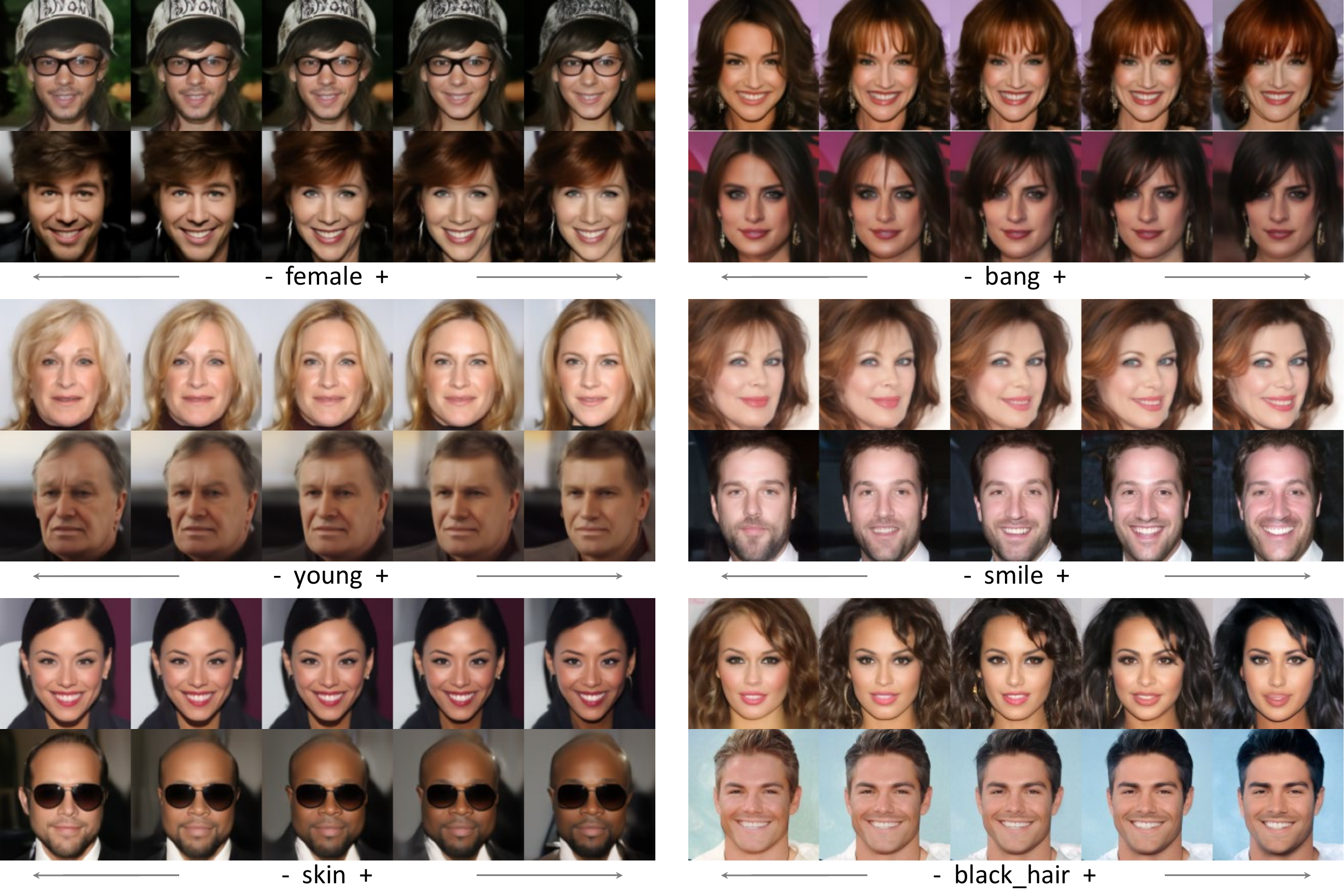}
    \vspace{-0.3cm}
    \caption{Interpretable directions discovered by "CelebAHQ128-500-40-256-2".}
    \label{fig:celebahq}
\end{figure*}

\subsection{Real Image Editing}
\begin{figure}[!h]
    \centering
    \includegraphics[width=0.45\textwidth]{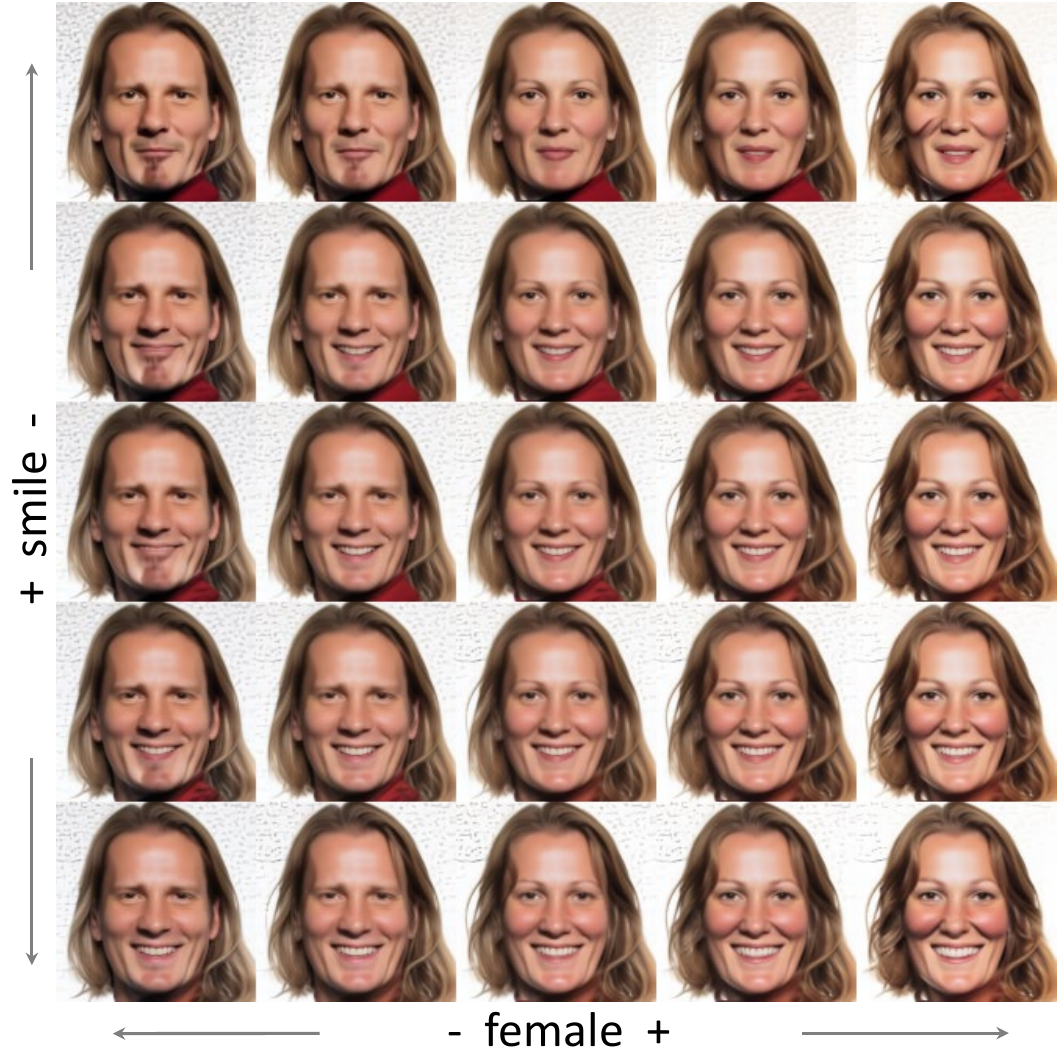}
    \caption{Real image editing examples. The center is the DDIM reconstruction of the real image. We edit it for the two interpretable directions discovered by "CelebAHQ128-500-40-256-2".}
    \label{fig:editing}
\end{figure}

In previous experiments, we always use Gaussian noise sampled from $\mathcal{N} (\mathbf{0}, \mathcal{I})$ to generate images, and we can extend our method to real image editing by generating from the corresponding noisy latent variable.
%  inverting any given real image through DDIM inversion.
% Thanks to DDIM inversion, we can achieve real image editing using pre-trained DMs.
Specifically, we randomly select an image from the FFHQ dataset~\cite{karras2019style} and invert it to $\widetilde{\mathbf{x}}_{T}$ through 100-step DDIM inversion using pre-trained DM on CelebAHQ.
Then we use "CelebAHQ128-500-40-256-2" to edit the image through Eq.(\ref{shift_reverse_process}) by starting from $\widetilde{\mathbf{x}}_{T}$ and using $\hat{\mathbf{h}}_{t} = \mathbf{h}_{t} + \sum_{i} s_{i} \cdot \Delta \mathbf{h}_{t}^{k_{i}}$, where $k_{i}$ is the $i$-th direction (attribute) we want to edit and $s_{i} \in [-S, S]$ is the corresponding editing strength.
Figure~\ref{fig:editing} show some editing examples for two editing directions.
As we can see, our method can smoothly transit the original image along the specified directions while maintaining other irrelevant attributes nearly stationary.
Moreover, the results further validate the linearity property of h-space.

\begin{figure*}[t]
    \centering
    \includegraphics[width=0.95\textwidth]{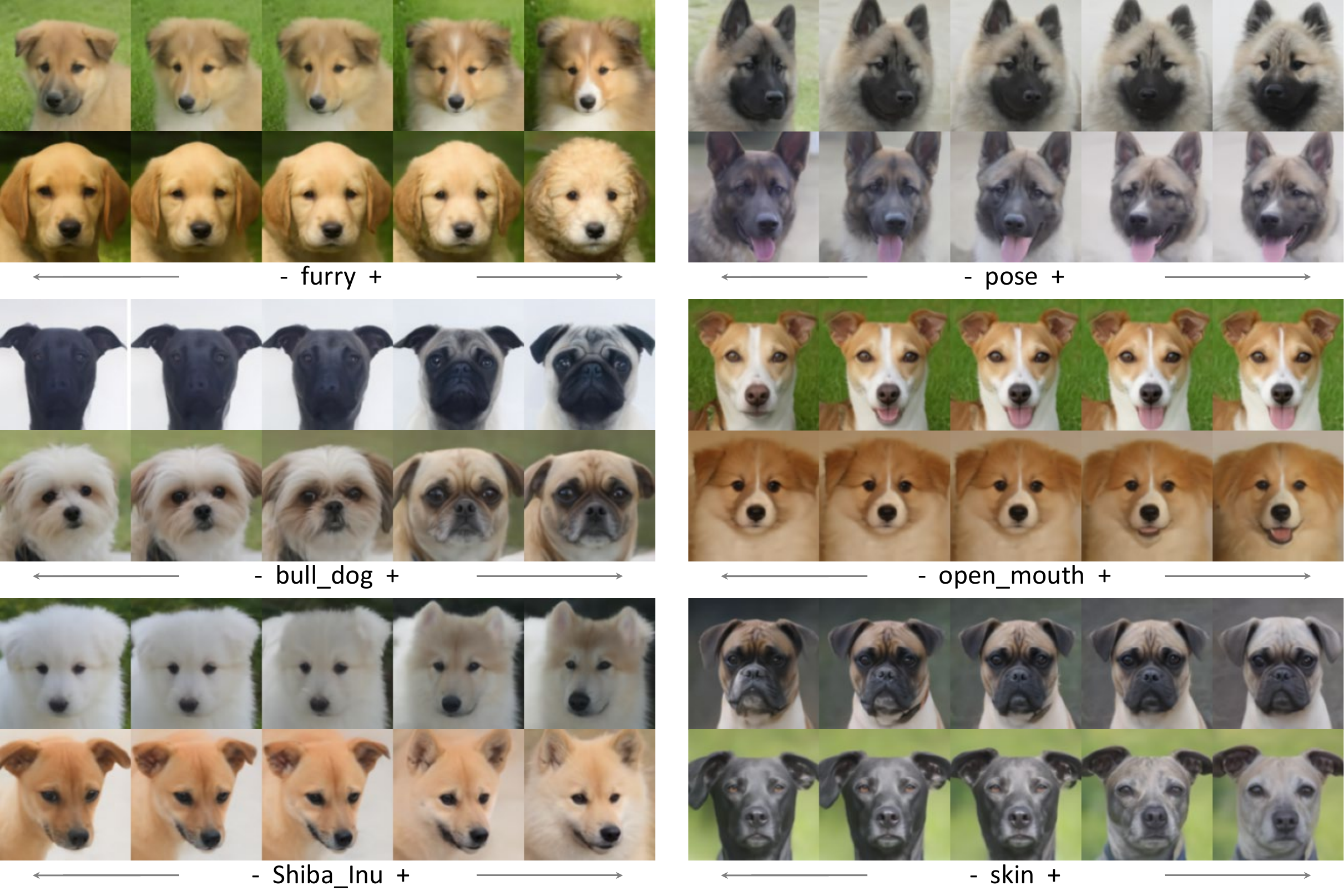}
    \vspace{-0.3cm}
    \caption{Interpretable directions discovered by "AFHQ128-400-40-256-2".}
    \label{fig:afhq}
\end{figure*}

\subsection{h-space Properties Validation}
\begin{figure}[t]
    \centering
    \includegraphics[width=0.45\textwidth]{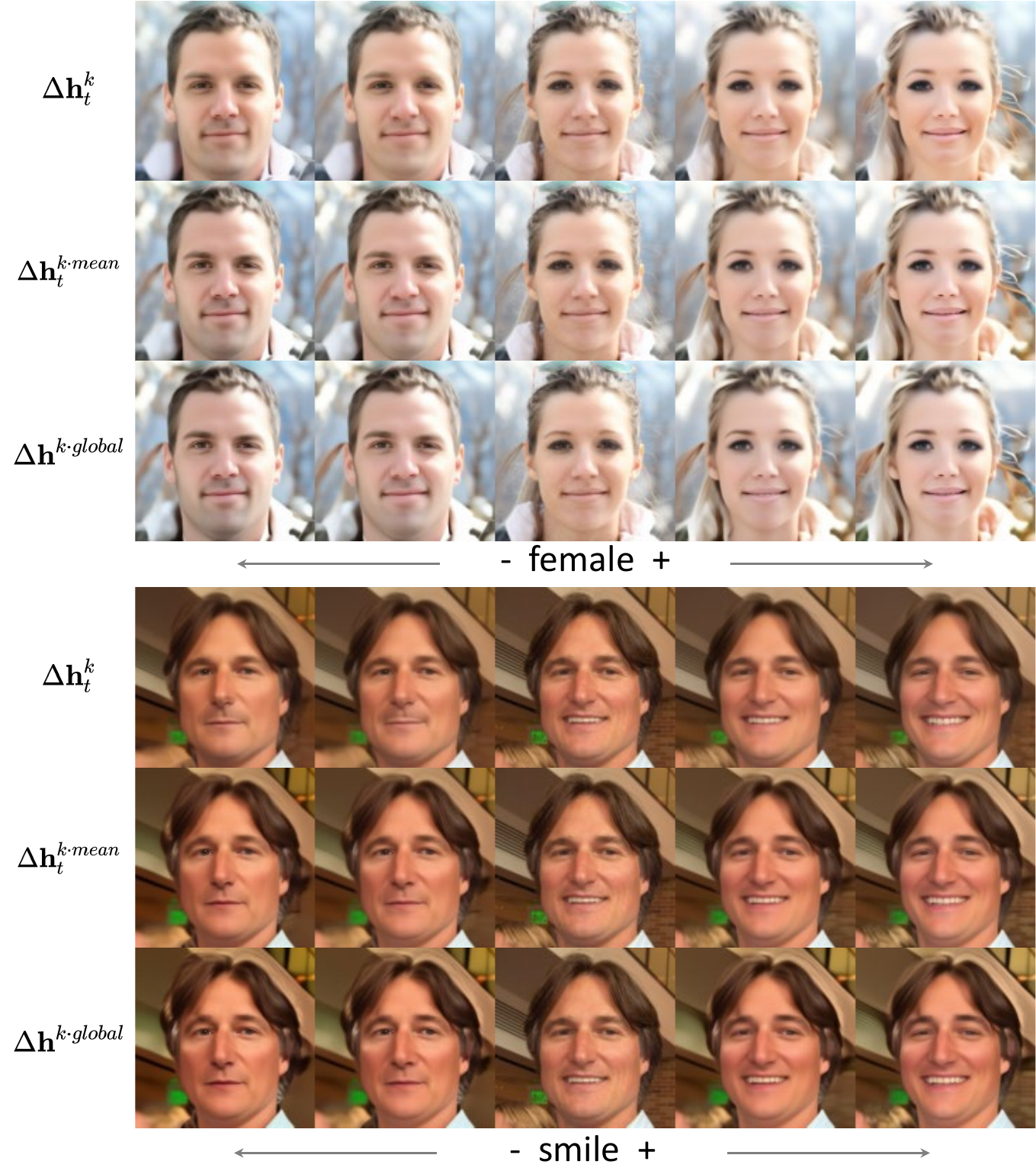}
    \caption{Real image editing examples using different directions. The middle of each row is the DDIM reconstruction of the real image.}
    \label{fig:property}
\end{figure}

Asyrp~\cite{kwon2022diffusion} has shown that h-space has several properties: homogeneity, linearity, and consistency across timesteps.
We have validated the linearity of our discovered semantic directions in previous experiments, and we now validate the remaining two.
Specifically, for a shift direction index $k$, we get the shift direction $\Delta \mathbf{h}_{t}^{k}$ of $20$ random samples and compute their mean direction as $\Delta \mathbf{h}_{t}^{k\cdot mean}$.
We also compute a time-invariant global direction as $\Delta \mathbf{h}^{k\cdot global} = \frac{1}{T_{s}} \sum_{t} \Delta \mathbf{h}_{t}^{k\cdot mean}$, where $T_{s}$ is the step number of shifting interval.
Then we respectively use $\hat{\mathbf{h}}_{t} = \mathbf{h}_{t} + \Delta \mathbf{h}_{t}^{k\cdot mean}$ and $\hat{\mathbf{h}}_{t} = \mathbf{h}_{t} + \Delta \mathbf{h}^{k\cdot global}$ for Eq.(\ref{shift_reverse_process}) to edit the real image selected from the FFHQ dataset.
Figure~\ref{fig:property} show two editing examples using different directions.
As we can see, all three directions almost produce the same effects for the real image, which validates the homogeneity and consistency of the directions our method discovers.

\subsection{Reconstructor Classification Accuracy (RCA) and Mean Opinion Score (MOS)}
We also evaluate our method in terms of quantitative results.
We follow GANLatentDiscovery~\cite{voynov2020unsupervised} to evaluate our method with RCA and MOS metrics.
Reconstructor Classification Accuracy (RCA) utilizes the trained reconstructor to evaluate the classification accuracy of shift directions given a batch of original samples and their corresponding shifted versions.
A high RCA indicates that the directions are easy to distinguish from each other, which implies that they decide different attributes of the samples.
For each model, we generate $5k$ pairs of samples with random shift direction indices and report their classification accuracy.
Mean Opinion Score (MOS) introduces a user study to quantify the interpretability of each individual direction.
We employ $16$ human assessors all having CV background for this study.
We prepare $8$ examples for each shift direction index $k \in \{1, K\}$, and for each example, we generate $9$ samples with $s$ uniformly varying from $[-S, S]$.
The assessors are asked to mark $1$ for each shift direction index $k$ only when $1)$ all $9$ samples of each example transit smoothly from $[-S, S]$, $2)$ the transition is consistent for all $8$ examples and $3)$ the transition is easy-to-interpret.
Otherwise it should be marked as $0$.
Each shift direction is assigned to $2$ different assessors.
We report the average of the marks across all assessors and shift directions.
Table~\ref{tab:quantitative} shows the results.
GANLatentDiscovery~\cite{voynov2020unsupervised} takes fixed random directions and directions corresponding to coordinate axes as baselines.
Unfortunately, we cannot follow these baselines because they will make the shifted sample out-of-domain.

\begin{table}[t]
    \centering
    \caption{Reconstructor Classification Accuracy and Mean Opinion Score of our models.}
    \label{tab:quantitative}
    \scalebox{0.93}{
        \begin{tabular}{cccc}
        \toprule
        \textbf{MNIST 32} & \textbf{Anime 64} & \textbf{CelebAHQ 128} & \textbf{AFHQ 128}  \\
        \midrule
        \multicolumn{4}{c}{\textbf{Reconstructor Classification Accuracy}} \\
        \midrule
        0.77 & 0.85 & 0.93 & 0.87 \\
        \midrule
        \multicolumn{4}{c}{\textbf{Mean Opinion Score}} \\
        \midrule
        0.52 & 0.36 & 0.28 & 0.31 \\
        \bottomrule
        \end{tabular}
    }
\end{table}

\subsection{Maintaining the Fidelity of Shifted Samples}
\begin{figure}[h]
    \centering
    \includegraphics[width=0.45\textwidth]{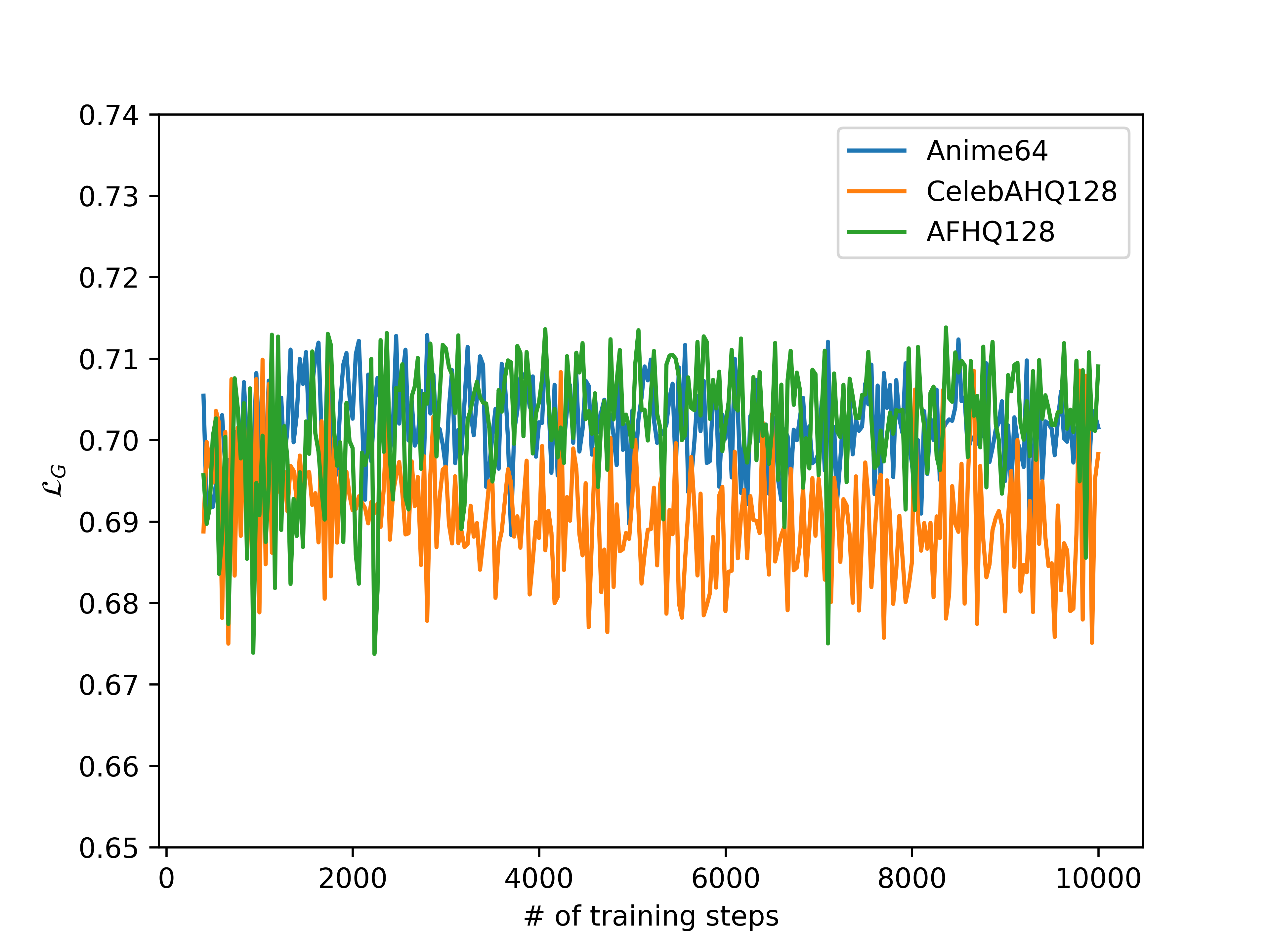}
    \caption{$\mathcal{L}_{G}$ of our models during training.}
    \label{fig:generator_loss}
\end{figure}

We state that our adversarial training for maintaining the fidelity of shifted samples reaches a stationary Nash equilibrium from the beginning and almost maintains it throughout.
We present the $\mathcal{L}_{G}$ of our models during training in Figure~\ref{fig:generator_loss}.
Evidently, $\mathcal{L}_{G}$ nearly sustains its ideal value of $- \log (0.5) \approx 0.6931$, thereby guaranteeing that our method identifies interpretable directions within the data distribution.

We further utilizes the FID scores to support this point.
Specifically, for each dataset, we generate raw samples using pre-trained DMs and shifted samples with random shift direction indices and magnitudes.
Then we calculate their FID to the real images randomly selected from dataset respectively.
Table~\ref{tab:fid} shows the results, which demonstrates that our discriminators can maintain the fidelity of shifted samples.

\begin{table}[h]
    \centering
    \caption{FID scores for pre-trained DMs (raw) and our models (shifted).}
    \label{tab:fid}
    \scalebox{0.9}{
        \begin{tabular}{l|ccc}
        \toprule
        & \textbf{Anime (50K)} & \textbf{CelebAHQ (30K)} & \textbf{AFHQ (4K)}  \\
        \midrule
        raw & 8.96 & 15.79 & 13.24 \\
        shifted & 9.21 & 16.70 & 12.65 \\
        \bottomrule
        \end{tabular}
    }
\end{table}

\section{Conclusion}
In conclusion, we present a general method to identify interpretable directions in the h-space of pre-trained diffusion models.
Our key idea is based on GANLatentDiscovery approach~\cite{voynov2020unsupervised} and the asymmetric DDIM reverse process~\cite{kwon2022diffusion}.
We also propose a general VRAM-efficient training algorithm to address the issue of excessive VRAM consumption caused by the iterative generative process of diffusion models.
Extensive experiments on various datasets demonstrate the effectiveness of our method.

% A primary limitation of our method is the relatively slow training and inference speed resulting from the multi-step iterative generative process of DMs.
% Although many studies have been able to achieve decent performance with few reverse steps, they still lag behind VAEs and GANs, which only need a single network pass.
% Another limitation is that we incorporate adversarial training to maintain the fidelity of shifted samples.
% Perhaps there exist other simpler methods to achieve this, and we leave their exploration as future work.

% The main potential negative impact of our work is the creation of deepfakes, involving the production of synthetic media that could be used for fraudulent, bullying, vengeful, or hoax purposes.
% Researchers have devised many algorithms similar to those used in building deepfakes to detect them.

{
    \small
    \bibliographystyle{ieeenat_fullname}
    \bibliography{reference}
}


\clearpage
\setcounter{page}{1}
\maketitlesupplementary
\section*{A. Algorithm}
Algorithm~\ref{algorithm:training} presents detailed procedure of VRAM-efficient algorithm combined with our method.

\begin{table*}[t]
    \centering
    \caption{Network architecture of pre-trained DMs.}
    \label{tab:pre-trained}
    \begin{tabular}{l|cccc}
    \toprule
    \textbf{Parameter} & \textbf{MNIST 32} & \textbf{Anime 64} & \textbf{CelebAHQ 128} & \textbf{AFHQ 128}  \\
    \midrule
    Base Channel & 64 & 64 & 128 & 128 \\
    Channel Multiplier & {[}1,2,2,4{]} & {[}1,2,4,8{]} & {[}1,1,2,3,4{]} & {[}1,1,2,3,4{]} \\
    ResBlock Num & 2 & 2 & 2 & 2 \\
    Attention Resolution(s) & None & 16 & 16 & 16 \\
    Attention Head Num & None & 4 & 4 & 4  \\
    Dropout & 0.0 & 0.0 & 0.1 & 0.1 \\
    % Images trained & 72M & 52M & 130M & 130M & 120M \\
    $\beta$ scheduler & \multicolumn{4}{c}{Linear} \\
    Training $T$ & \multicolumn{4}{c}{1000} \\
    Diffusion Loss & \multicolumn{4}{c}{MSE with noise prediction $\epsilon$} \\   
    \bottomrule
    \end{tabular}
\end{table*}

\begin{table*}[t]
    \centering
    \caption{Training configurations of our models.}
    \label{tab:training}
    \begin{tabular}{l|cccc}
    \toprule
    \textbf{Parameter} & \textbf{MNIST 32} & \textbf{Anime 64} & \textbf{CelebAHQ 128} & \textbf{AFHQ 128}  \\
    \midrule
    K & 32 & 128 & 256 & 256 \\
    S & 5 & 2 & 2 & 2 \\
    M & 20 & 40 & 40 & 40 \\
    $t_{\text{stop}}$ & 400 & 400 & 500 & 400 \\
    $\lambda_{1}$ & \multicolumn{4}{c}{0.1} \\
    $\lambda_{2}$ & \multicolumn{4}{c}{0.1} \\
    Optimizer & \multicolumn{4}{c}{Adam with default parameters}  \\
    Learning Rate & \multicolumn{4}{c}{1e-3} \\
    Batch Size & 128 & 128 & 64 & 64 \\
    Training Steps & 3k & 12k & 15k & 15k \\
    \bottomrule
    \end{tabular}
\end{table*}

\section*{B. Implementation Details}
\subsection*{B.1 Pre-trained Diffusion Models}
We use the ADM codebase~\cite{dhariwal2021diffusion} in~\href{https://github.com/openai/guided-diffusion}{guided-diffusion} to train DMs on four datasets.
The configuration of these pre-trained DMs that we employ can be found in Table~\ref{tab:pre-trained}.

\subsection*{B.2 Reconstructor}
For reconstructor $R^{\omega}$, we follow GANLatentDiscovery~\cite{voynov2020unsupervised} to use the LeNet~\cite{lecun1998gradient} for MNIST and AnimeFaces and the ResNet-18 model~\cite{he2016deep} for CelebAHQ and AFHQ-dog.

\subsection*{B.3 Discriminator}
We follow~\cite{kim2022refining} to use the encoder part of UNet~\cite{ronneberger2015u} followed by an AdaptiveAvgPool layer as our discriminator $D^{\psi}$.
The UNet encoder configuration is the same as that of the corresponding pre-trained DM, as presented in Table~\ref{tab:pre-trained}, so that the discriminator is completely determined by pre-trained DPMs and can be universally applied to different UNet architectures.

\subsection*{B.4 Training Details}
We show our training configurations in Table~\ref{tab:training}.
To determine the editing interval, Asyrp~\cite{kwon2022diffusion} proposes a strategy to seek the shortest (i.e., infimum) editing interval which will bring enough distinguishable changes in the images in general.
Therefore, the editing interval varies for different editing targets.
In view of minor differences among different editing targets for the same dataset, we opt to use their average as our shifting interval, which can satisfy the identification of most potential interpretable directions.
All the experiments are performed on $8$ Nvidia RTX 3090 GPUs.

\section*{C. Additional Samples}
Figure~\ref{fig:anime}~\ref{fig:celebahq}~\ref{fig:afhq} show more interpretable directions discovered by "Anime64-400-40-128-2", "CelebAHQ128-500-40-256-2" and "AFHQ128-400-40-256-2", respectively.

\section*{D. Limitations and Broader Impacts}
A primary limitation of our method is the relatively slow training and inference speed resulting from the multi-step iterative generative process of DMs.
Although many studies have been able to achieve decent performance with few reverse steps, they still lag behind VAEs and GANs, which only need a single network pass.
Another limitation is that we incorporate adversarial training to maintain the fidelity of shifted samples.
Perhaps there exist other simpler methods to achieve this, and we leave their exploration as future work.

The main potential negative impact of our work is the creation of deepfakes, involving the production of synthetic media that could be used for fraudulent, bullying, vengeful, or hoax purposes.
Researchers have devised many algorithms similar to those used in building deepfakes to detect them.

\begin{figure*}[t]
    \centering
    \includegraphics[width=1.0\textwidth]{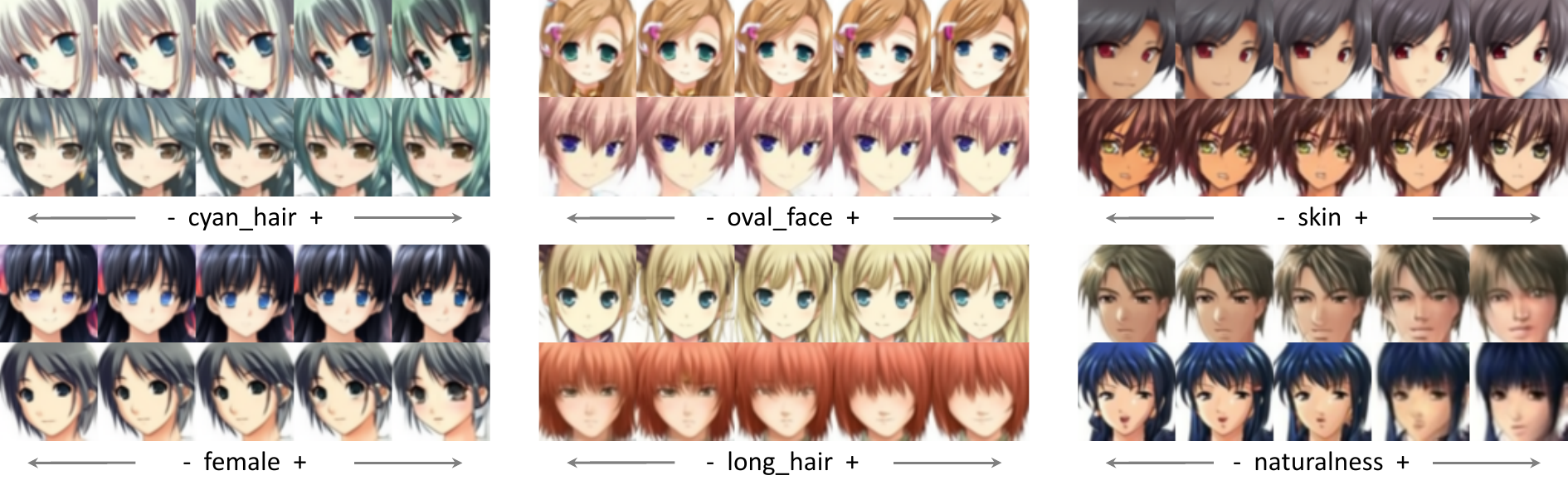}
    \caption{Interpretable directions discovered by "Anime64-400-40-128-2".}
    \label{fig:anime}
\end{figure*}

\begin{figure*}[t]
    \centering
    \includegraphics[width=1.0\textwidth]{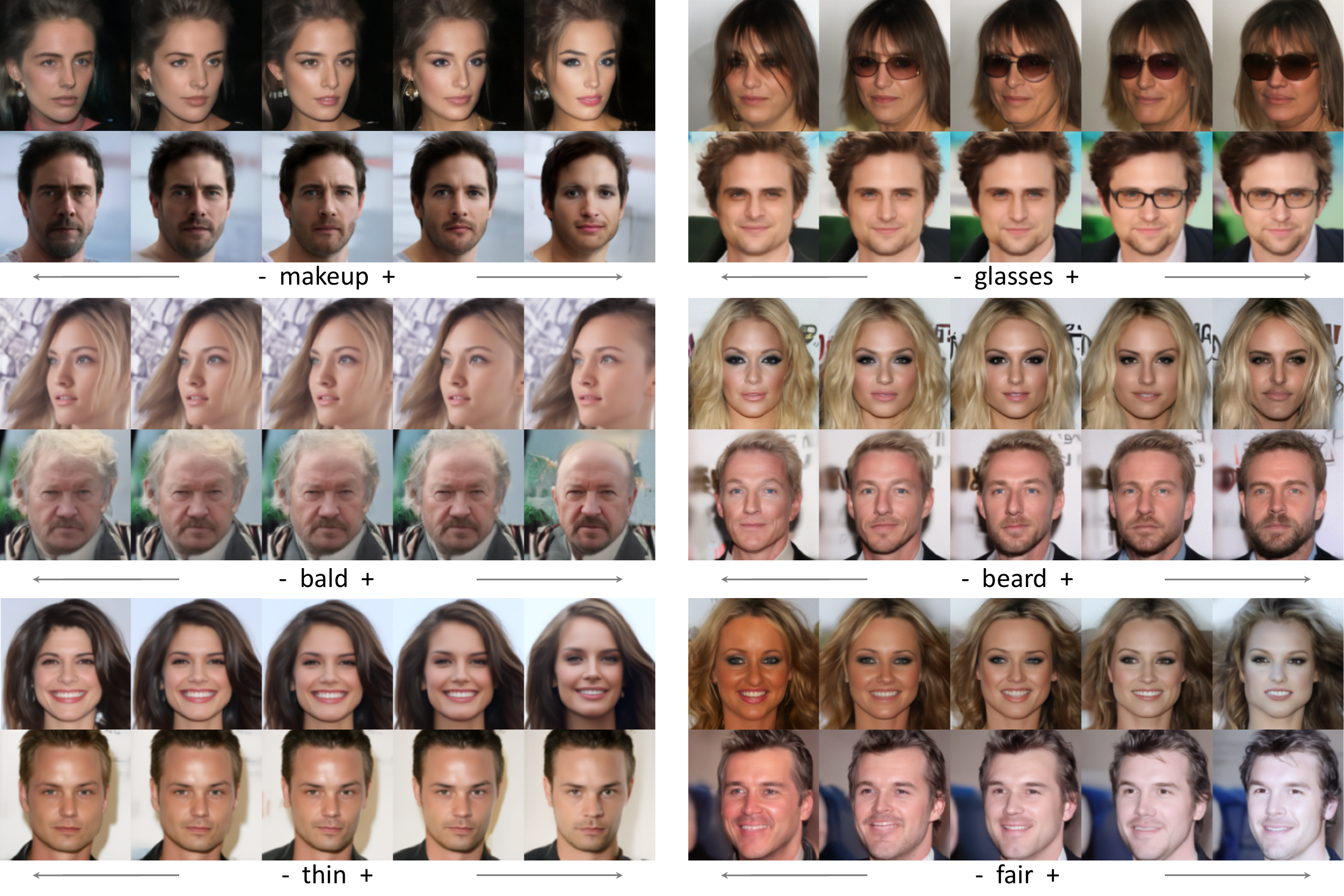}
    \caption{Interpretable directions discovered by "CelebAHQ128-500-40-256-2".}
    \label{fig:celebahq}
\end{figure*}

\begin{figure*}[t]
    \centering
    \includegraphics[width=1.0\textwidth]{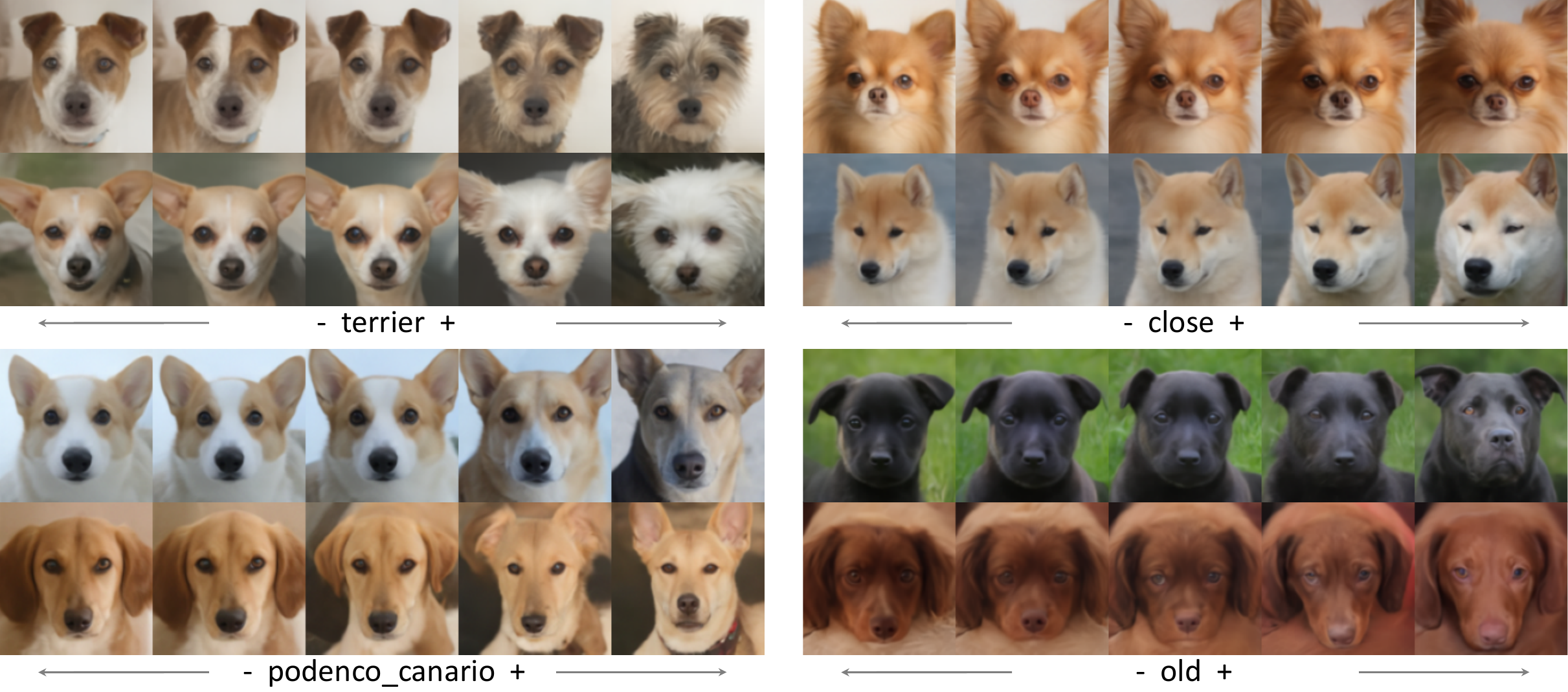}
    \caption{Interpretable directions discovered by "AFHQ128-400-40-256-2".}
    \label{fig:afhq}
\end{figure*}

\begin{algorithm*}[t] 
    \SetKwBlock{DummyBlock}{\PyCode{with torch.no\underline{~}grad():}}{}
    \caption{PyTorch-Style VRAM-efficient Training Algorithm Pseudocode\label{algorithm:training}}
    \KwIn{$\epsilon_{t}^{\theta}$ (pre-trained diffusion model), $K$ (\# of directions), $S$ (max magnitude multiplied to $\Delta \mathbf{h}_{t}^{k}$), $M$(\# of generation steps), $\{ \tau_{i}\}^{M}_{i=1}$ (generation timestep sequence  where $\tau_{1}=0$ and $\tau_{M}=T$), $t_{\textnormal{stop}}$ (shifting stop timestep), $\lambda_{1},\lambda_{2}$ (weights of different losses)}
    % $\gamma$ (smoothing constant for $\mathcal{L}_{1}$ loss),

    \PyCode{$\mathbf{x}_{T} \sim \mathcal{N} (\mathbf{0}, \mathcal{I})$\textnormal{;} $\widetilde{\mathbf{x}}_{T}$ = $\mathbf{x}_{T}$\textnormal{;} k$\sim\mathcal{U}\{1, K\}$\textnormal{;} s$\sim\mathcal{U}[-S, S]$\textnormal{;} node\underline{~}list=[$\widetilde{\mathbf{x}}_{T}$]} \\
    % \textnormal{;} \PyCode{P\underline{~}list=$[]$}

    \PyComment{step 1: train discriminator and record nodes} \\
    \SetAlgoNoLine\DummyBlock{\SetAlgoVlined
        \For{$\ i = M, M-1, \cdots, 2$}
        {
            \PyComment{generate real sample} \\
            $\mathbf{P}^{r}_{\tau_{i}} = \mathbf{P}_{\tau_{i}}\big(\epsilon_{\tau_{i}}^{\theta}(\mathbf{x}_{\tau_{i}})\big)$\textnormal{;} $\mathbf{D}^{r}_{\tau_{i}} = \mathbf{D}_{\tau_{i}}\big(\epsilon_{\tau_{i}}^{\theta}(\mathbf{x}_{\tau_{i}})\big)$\textnormal{;} $\mathbf{x}_{\tau_{i-1}} = \sqrt{\bar{\alpha}_{\tau_{i-1}}} \mathbf{P}^{r}_{\tau_{i}} + \mathbf{D}^{r}_{\tau_{i}}$ \\

            \PyComment{generate fake sample and record nodes} \\
            \uIf{$\tau_{i} \ge t_{\textnormal{stop}}$}{
                % \PyComment{Asymmetric reverse process} \\
                Extract $\mathbf{h}_{\tau_{i}}$ from $\epsilon_{\tau_{i}}^{\theta}(\widetilde{\mathbf{x}}_{\tau_{i}})$\textnormal{;} $\Delta \mathbf{h}_{\tau_{i}}^{k} = f_{\tau_{i}}^{\phi}(\mathbf{h}_{\tau_{i}}, k)$\textnormal{;} $\hat{\mathbf{h}}_{\tau_{i}} = \mathbf{h}_{\tau_{i}} + s \cdot \Delta \mathbf{h}_{\tau_{i}}^{k}$\\
                $\mathbf{P}^{f}_{\tau_{i}} = \mathbf{P}_{\tau_{i}}\big(\epsilon_{\tau_{i}}^{\theta}(\widetilde{\mathbf{x}}_{\tau_{i}} | \hat{\mathbf{h}}_{\tau_{i}})\big)$\textnormal{;} $\mathbf{D}^{f}_{\tau_{i}}=\mathbf{D}_{\tau_{i}}\big(\epsilon_{\tau_{i}}^{\theta}(\widetilde{\mathbf{x}}_{\tau_{i}} | \mathbf{h}_{\tau_{i}} )\big)$\textnormal{;} $\widetilde{\mathbf{x}}_{\tau_{i-1}} = \sqrt{\bar{\alpha}_{\tau_{i-1}}}\mathbf{P}^{f}_{\tau_{i}} + \mathbf{D}^{f}_{\tau_{i}}$
            }
            \Else{
                % \PyComment{DDIM reverse process} \\
                $\mathbf{P}^{f}_{\tau_{i}} = \mathbf{P}_{\tau_{i}}\big(\epsilon_{\tau_{i}}^{\theta}(\widetilde{\mathbf{x}}_{\tau_{i}} | \mathbf{h}_{\tau_{i}} )\big)$\textnormal{;} $\mathbf{D}^{f}_{\tau_{i}} = \mathbf{D}_{\tau_{i}}\big(\epsilon_{\tau_{i}}^{\theta}(\widetilde{\mathbf{x}}_{\tau_{i}} | \mathbf{h}_{\tau_{i}} )\big)$\textnormal{;} $\widetilde{\mathbf{x}}_{\tau_{i-1}} = \sqrt{\bar{\alpha}_{\tau_{i-1}}}\mathbf{P}^{f}_{\tau_{i}} + \mathbf{D}^{f}_{\tau_{i}}$
            } 
        
            % \uIf{$\tau_{i} \ge t_{\textnormal{stop}}$}{
            %     \PyCode{$\mathcal{L}^{D}_{\tau_{i}}$ = nn.BCELoss($D^{\psi}(\mathbf{P}^{r}_{\tau_{i}}, \tau_{i}), 1$) + nn.BCELoss($D^{\psi}(\mathbf{P}^{f}_{\tau_{i}}, \tau_{i}), 0$)} \\
            %     \PyCode{$\mathcal{L}^{D}_{\tau_{i}}$.backward()} \PyComment{compute and accumulate grad to discriminator} \\ 
            % }

            % \PyCode{P\underline{~}list.append($\mathbf{P}^{r}_{\tau_{i}}$)} \\
            \uIf{$i > 2$}{
                \PyCode{node\underline{~}list.append($\widetilde{\mathbf{x}}_{\tau_{i-1}}$)} \PyComment{record nodes}
            }
        }
    }\SetAlgoVlined
    
    \PyCode{$\mathcal{L}_{D}$ = nn.BCELoss($D^{\psi}(\mathbf{x}_{0}), 1$) + nn.BCELoss($D^{\psi}(\widetilde{\mathbf{x}}_{0}), 0$)}\textnormal{;} \PyCode{$\mathcal{L}_{D}$.backward()} \\
    Take a gradient step for $\psi$  \PyComment{update discriminator} \\
    \PyComment{step 2: train shift block and reconstructor} \\
    \For{$\ i = 2, 3, \cdots, M $}{
        \PyCode{node = node\underline{~}list.pop()\textnormal{;} node.requires\underline{~}grad\underline{~}(True)} \\
        \uIf{$\tau_{i} < t_{\textnormal{stop}}$}{
            $\mathbf{P}_{\tau_{i}} = \mathbf{P}_{\tau_{i}}\big(\epsilon_{\tau_{i}}^{\theta}($node$ | \mathbf{h}_{\tau_{i}} )\big)$\textnormal{;} $\mathbf{D}_{\tau_{i}} = \mathbf{D}_{\tau_{i}}\big(\epsilon_{\tau_{i}}^{\theta}($node$ | \mathbf{h}_{\tau_{i}} )\big)$\textnormal{;} $\widetilde{\mathbf{x}}_{\tau_{i-1}} = \sqrt{\bar{\alpha}_{\tau_{i-1}}}\mathbf{P}_{\tau_{i}} + \mathbf{D}_{\tau_{i}}$
        }
        \Else{
            Extract $\mathbf{h}_{\tau_{i}}$ from $\epsilon_{\tau_{i}}^{\theta}($node$)$\textnormal{;} $\Delta \mathbf{h}_{\tau_{i}}^{k} = f_{\tau_{i}}^{\phi}(\mathbf{h}_{\tau_{i}}, k)$\textnormal{;} $\hat{\mathbf{h}}_{\tau_{i}} = \mathbf{h}_{\tau_{i}} + s \cdot \Delta \mathbf{h}_{\tau_{i}}^{k}$ \\
            $\mathbf{P}_{\tau_{i}} = \mathbf{P}_{\tau_{i}}\big(\epsilon_{\tau_{i}}^{\theta}($node$ | \hat{\mathbf{h}}_{\tau_{i}})\big)$\textnormal{;} $\mathbf{D}_{\tau_{i}}=\mathbf{D}_{\tau_{i}}\big(\epsilon_{\tau_{i}}^{\theta}($node$ | \mathbf{h}_{\tau_{i}} )\big)$\textnormal{;} $\widetilde{\mathbf{x}}_{\tau_{i-1}} = \sqrt{\bar{\alpha}_{\tau_{i-1}}}\mathbf{P}_{\tau_{i}} + \mathbf{D}_{\tau_{i}}$
        }

        \uIf{$i == 2$}{
            % \PyCode{$\mathcal{L}_{1}$ = $\lambda_{3}\cdot\frac{\gamma}{\gamma+|s|}\cdot$ nn.L1Loss($\widetilde{\mathbf{x}}_{0}, \mathbf{x}_{0}$)} \PyComment{$\widetilde{\mathbf{x}}_{0}$ is just $\widetilde{\mathbf{x}}_{\tau_{i-1}}$, $\mathbf{x}_{0}$ is from step 1} \\
            \PyCode{$\mathcal{L}_{G}$ = nn.BCELoss($D^{\psi}(\widetilde{\mathbf{x}}_{0}), 1$)} \PyComment{$\widetilde{\mathbf{x}}_{0}$ is just $\widetilde{\mathbf{x}}_{\tau_{i-1}}$} \\
            \PyCode{logit, shift = $R^{\omega}(\widetilde{\mathbf{x}}_{0}, \mathbf{x}_{0})$} \PyComment{$\hat{k}$ and $\hat{s}$, $\mathbf{x}_{0}$ is from step 1} \\
            \PyCode{$\mathcal{L}_{R}$ = $\lambda_{1}\cdot$nn.CrossEntropyLoss(logit, k) + $\lambda_{2}\cdot$nn.L1Loss(shift, s)} \\
            \PyCode{($\mathcal{L}_{G}$ + $\mathcal{L}_{R}$).backward()} \PyComment{accumulate grad to reconstructor} \\ 
            \PyCode{node\underline{~}grad = node.grad.clone()} \\
        }
        \ElseIf{$\tau_{i} < t_{\textnormal{stop}}$}{
            \PyCode{node\underline{~}grad = autograd.grad($\widetilde{\mathbf{x}}_{\tau_{i-1}}$, node, grad\underline{~}outputs=node\underline{~}grad)}
        }
        \Else{
            \PyCode{$\widetilde{\mathbf{x}}_{\tau_{i-1}}$.backward(node\underline{~}grad)} \PyComment{accumulate grad to shift block} \\
            \PyCode{node\underline{~}grad = node.grad.clone()} \\
        }
    }
    Take a gradient step for $\phi$ and $\omega$ \PyComment{update shift block and reconstructor} \\
\end{algorithm*}

\end{document}